\documentclass{article}

\usepackage[preprint]{neurips_2022}

\usepackage[utf8]{inputenc} 
\usepackage[T1]{fontenc}    
\usepackage{hyperref}       
\usepackage{url}            
\usepackage{booktabs}       
\usepackage{amsfonts}       
\usepackage{nicefrac}       
\usepackage{microtype}      
\usepackage{xcolor}         
\usepackage{booktabs} 
\usepackage{threeparttable, tablefootnote}
\usepackage{algorithm}
\usepackage{algpseudocode}
\usepackage{amsmath,amsfonts,amsthm,bm, bbm}
\usepackage{diagbox}
\usepackage{amsmath, bm, bbm}
\usepackage{amssymb}
\usepackage{mathtools}
\usepackage{amsthm}
\usepackage{adjustbox}
\usepackage{multirow}

\usepackage{xcolor}

\usepackage{pifont}
\newcommand{\xmark}{\text{\ding{55}}}

\DeclareMathOperator*{\argmax}{arg\,max}
\DeclareMathOperator*{\argmin}{arg\,min}

\title{Key Design Choices for Double-Transfer in Source-Free Unsupervised Domain Adaptation}

\author{%
  Andrea Maracani$^{1, 2, 3}$, Raffaello Camoriano$^{5*}$, Elisa Maiettini$^{1*}$, Davide Talon$^{1, 2}$, \\
  \textbf{Lorenzo Rosasco}$^{2, 3, 4}$, \textbf{Lorenzo Natale}$^1$\\
  $^1$Istituto Italiano di Tecnologia (IIT)\\
  $^2$Università degli Studi di Genova\\
  $^3$DIBRIS \& MaLGA\\
  $^4$IIT, CBMM - MIT\\
 Genoa, Italy \\
  $^5$Politecnico di Torino\\
  Turin, Italy \\
  \{\texttt{name.surname}\}\texttt{@iit.it}, \texttt{raffaello.camoriano@polito.it}, 
  \texttt{lrosasco@mit.edu}
}
\begin{document}

\maketitle
\def\thefootnote{*}\footnotetext{These authors contributed equally.}\def\thefootnote{\arabic{footnote}}

\begin{abstract}
Fine-tuning and Domain Adaptation emerged as effective strategies for efficiently transferring deep learning models to new target tasks. However, target domain labels are not accessible in many real-world scenarios. This led to the development of Unsupervised Domain Adaptation (UDA) methods, which only employ unlabeled target samples. Furthermore, efficiency and privacy requirements may also prevent the use of source domain data during the adaptation stage. This  challenging setting, known as Source-Free Unsupervised Domain Adaptation (SF-UDA), is gaining interest among researchers and practitioners due to its potential for real-world applications.\\
In this paper, we  provide the first in-depth analysis of the main design choices in SF-UDA through a large-scale empirical study across 500 models and 74 domain pairs. 
We pinpoint the normalization approach, pre-training strategy, and backbone architecture as the most critical factors. 
Based on our  quantitative findings, we propose recipes to best tackle SF-UDA scenarios. 
Moreover, we show that SF-UDA is competitive also beyond standard benchmarks and backbone architectures, performing on par with UDA at a fraction of the data and computational cost. 
In the interest of reproducibility, we include the full experimental results and code as supplementary material.
\end{abstract}

\section{Introduction}
The recent success of deep neural networks (DNNs) in many tasks and domains often relies on the availability of large annotated datasets.
This can be tackled by pre-training DNNs on a large dataset and then fine-tuning their weights with target task data~\citep{huh2016makes, yosinski2014transferable, chu2016best}.
Furthermore, fine-tuning (FT) is usually simpler and faster than training the model from scratch, the dataset size can be smaller, and the final performance is typically higher (with some exceptions: see the work by \cite{kornblith2019better}). 
This approach is very convenient: the model requires a single expensive pre-training and can later be re-used for multiple downstream tasks.
This is a good example of transfer learning~\citep{Zhuang2021Comprehensive}, which
leverages on the information acquired from a  task to improve accuracy on another task of interest.
Two relevant examples of transfer learning are Domain Adaptation (DA), that, given different--yet related--tasks, exploits source domain(s) data to improve performance on different known target domain(s), and Domain Generalization (DG), that aims to generalize to unknown target(s).

As opposed to fine-tuning, in which the pre-training and downstream tasks can be significantly different, DA and DG require stronger assumptions on the similarity between tasks, e.g., leveraging synthetic images to improve the classification of real images that share the same label space.
DA is also related to Multi-Task Learning (MTL)~\citep{caruana1997multitask,ciliberto2017consistent} and Multi-Domain Learning (MDL)~\citep{joshi2012multi}. 
In fact, domains can be seen as tasks in MTL or MDL. Still, an explicit domain label is provided  and annotations are available for each task.

A particularly challenging and useful setting in practice is Unsupervised Domain Adaptation 
(UDA)~\citep{tzeng2017adversarial, ganin2016domain}, in which labeled samples from a source domain are used together with unlabeled samples from the target domain to improve performance on the latter.
This work focuses on Source-Free Unsupervised Domain Adaptation (SF-UDA)~\citep{liang2020we} for the image classification task. SF-UDA is a two-steps sequential version of UDA in which the source-domain labeled data is only accessible in the first training phase. 
Adaptation to the new domain is carried out in a second stage where only the unlabeled data from the target domain is available. 

SF-UDA nicely matches applications where adaptation is required with computational and memory constraints, 
or where privacy policies prevent access to the source data. 
Since these techniques usually fine-tune a model (with pre-trained weights) to the source domain and then they adapt it to the target domain, there are two different transfers into play: (1) from the base task (used for pre-training) to the source domain, and (2) from the source domain to the target domain: we refer to this combined transfer as \textit{double-transfer}.
The motivation of this work lies in the typical questions that arise when facing a new SF-UDA task to solve. 
We note that a systematic study and best practices on how to tackle double-transfer in SF-UDA are currently lacking in the literature.
Firstly, we experimentally identify the main design choices in terms of their impact on downstream performance: we find out that the backbone architecture, pre-training dataset, and the way double-transfer is performed play a critical role. 
Hence, we rigorously quantify and analyze the impact of each factor.
Secondly, we empirically investigate the strengths and failure modes of SF-UDA methods, also comparing them with UDA ones.

Recent works~\citep{liang2020we, ding2022source} show that SF-UDA techniques achieve comparable performance with state-of-the-art UDA methods on common benchmarks. 
In contrast, \cite{kim2022broad} reports that recent UDA methods perform well on standard benchmarks because they \textit{overfit} the task. 
Indeed, when employed in other settings (i.e., non-standard architectures or datasets) they result in worse accuracy than previous methods. 
We investigate with targeted experiments whether overfitting affects recent SF-UDA methods as well.

We pursue such objectives through large-scale systematic experiments encompassing more than 500 different architectures on 6 separate domain adaptation datasets, totaling 23 domains and 74 domain shifts. 
We employ different probing and SF-UDA methods to better analyze the functioning of double-transfer methods, providing a ready-to-use recipe for effective system design. Our main findings are as follows:

\begin{itemize}
    \item The pre-training dataset choice and the resulting accuracy on the ImageNet top-1 benchmark directly impact the domain generalization and SF-UDA performance, for both CNNs and Vision Transformers. Switching from ImageNet-1K to ImageNet-21K boosts the average performance up to $\sim10\%$, see Sec.~\ref{modselection}.

     \item Most SF-UDA methods fine-tune the model on the source domain before adaptation. 
     However, we show that in some cases this causes severe performance degradation. Specifically, we identify the type of normalization layers as having a critical role in this context. On average, fine-tuning causes a performance degradation of 4.6\% for models with Batch Normalization (BN)~\citep{ioffe2015batch}, while models with Layer Normalization (LN)~\citep{ba2016layer} benefit from it, see Sec.~\ref{ft_sec}.

     \item Besides fine-tuning, the normalization strategy heavily affects  the failure rate of SF-UDA in general. 
     We present a large-scale analysis of SF-UDA methods' failure rates, comparing architectures with LN and BN and attesting the improved robustness of the former. On average, LN has a lower failure rate than BN, by a margin up to $\sim11\%$, see Sec.~\ref{robust_sec}.
   
    \item SF-UDA methods, like SHOT~\citep{liang2020we}, SCA (see Sec.~\ref{sf_uda}), and NRC~\citep{yang2021exploiting}, perform well also with architectures and datasets different from the usual benchmark ones and are competitive with state-of-the-art UDA methods, see Sec.~\ref{sec:vsuda}. 
\end{itemize}

The full list of architectures, results in csv format, the code and pre-trained weights are released in the supplementary material.

\section{Background and Related Work}
\label{related_sec}

Let $\mathcal{X}$ be the input space, e.g., the image  space, $\mathcal{Z} \subseteq \mathbb{R}^D$ a representation space, i.e., the feature space, and $\mathcal{Y} = \{1, \ldots, C\}$ the output space for multi-class classification.
A \textit{feature extractor} (backbone) is a function $f_{\bm{\theta}}: \mathcal{X} \to \mathcal{Z}$ with parameter $\bm{\theta}$, while a \textit{classifier} is a function with parameter $\bm{\phi}$ that assigns a label to any feature vector, $h_{\bm{\phi}}: \mathcal{Z} \to \mathcal{Y}$.
We introduce two data distributions over $\mathcal{X} \times \mathcal{Y}$: $\mu_S$ that models the source domain and $\mu_T$ that models the target domain.\\
\textbf{Domain Generalization (DG}). In this work, we consider DG setting as the reference task to investigate transferability among domains. First introduced by \cite{blanchard2011generalizing}, its goal is to find a feature extractor $f_{\bm{\theta}}$ and a classifier $h_{\bm{\phi}}$ from $N$ \textit{i.i.d.} samples of a given domain (the source $\mu_S$) that will perform well on other unseen domains (in our case just the target domain $\mu_T$), that is:

\vspace{-0.4cm}
\begin{align}
&\min_{\bm{\theta}, \bm{\phi}} \mathbb{E}_{(x, y) \sim \mu_T}[\mathbbm{1}\{h_{\bm{\phi}}(f_{\bm{\theta}}(x)) \neq y\}]  \\
&\text{given } (x_i, y_i)_{i=1}^N \sim \mu_S^N,
\end{align}
\vspace{-0.3cm}

where $\mathbbm{1}\{\text{condition}\}$ is the indicator function. 
In this setting, some assumptions on the relationship between tasks ($\mu_S$ and $\mu_T$) are needed, but only data sampled from the source domain can be used for training, while target domain data is accessible at test time only. 
We remark that this is similar to the standard supervised-learning problem of generalization (Gen), with the exception that, here, the training and test distributions are different. 
Several methods specifically target this problem \citep{volpi2018generalizing, arjovsky2019invariant, ilse2020diva}; see the work of \cite{ijcai2021-628} for a review.\\
\textbf{Usupervised Domain Adaptation (UDA).} As for DG, the final goal of UDA is to learn a model that performs well on the target domain. However, differently from DG, more information is available: unlabeled samples from the target (marginal) distribution are accessible together with the labeled data from the source. 
Since the seminal theoretical works \cite{ben2006analysis, ben2010theory, mansour2009domain}, many UDA methods have been proposed in the literature for image classification, including adversarial training~\citep{ganin2016domain}, bidirectional matching~\citep{na2021fixbi}, per-class kernel mean discrepancy minimization~\citep{kang2019contrastive}, and also for other tasks such as object detection~\citep{oza2021unsupervised} and semantic segmentation~\citep{toldo2020unsupervised}.\\
\textbf{Source-Free Unsupervised Domain Adaptation (SF-UDA).} SF-UDA is similar to UDA, but it is more constrained.
The learning process is divided into 2 phases: the (labeled) source data is available only in the first training step.
Then, adaptation to the new domain occurs in the second stage, where only the unlabeled target data is available. The problem can be formulated as:

\begin{align}
& \min_{\bm{\theta}, \bm{\phi}} \mathbb{E}_{(x, y) \sim \mu_T}[\mathbbm{1}\{h_{\bm{\phi}}(f_{\bm{\theta}}(x)) \neq y\}] \\
& \text{given } (x_i, y_i)_{i=1}^N \sim \mu_S^N \text{ (only at step 1)} \\
& \text{and } (x_j)_{j=1}^M \sim \mu_{T(\mathcal{X})}^M \text{ (only at step 2)} 
\end{align}

where $\mu_{T(\mathcal{X})}$ is the marginal of $\mu_T$ over the input space $\mathcal{X}$ and $M$ is the number of available target samples. We remark that, even if it is not possible to share data among the 2 steps, model parameters $\bm{\theta}'$ and $\bm{\phi}'$ found at step 1 are accessible at step 2. 
Recently, SF-UDA has gained significant interest: since the work of \cite{liang2020we}, many other methods have been proposed and they achieved remarkable results on standard UDA benchmarks~\citep{li2020model, yang2021generalized, kundu2020universal, huang2021model, kurmi2021domain, xia2021adaptive, chu2022denoised, ding2022source, kundu2022balancing}. 
SF-UDA methods have also been recently applied to Natural Language Processing tasks~\citep{su2022comparison}.
In Sec.~\ref{sf_uda}, we will illustrate some SF-UDA methods in greater detail.\\
\textbf{Experimental Studies.} Training DNNs is demanding in terms of computational resources, time and data. 
Hence, there is great interest in the scientific community into understanding how to obtain representations that can be 
conveniently
transferred to new tasks and what are the key ingredients to build more efficient architectures and training methods.
For this reason, several studies  on transfer learning (i.e., fine-tuning for classification or different vision tasks) have been conducted, such as the works of \cite{chu2016best, huh2016makes, kornblith2019better}.
Concerning UDA, the work of \cite{zhang2020impact} studies model selection for classical methods (and for CNNs) based on the  accuracy achieved on ImageNet, while \cite{kim2022broad} provides an analysis of different pre-training techniques for DG and UDA. 
In this work, instead, we target SF-UDA approaches. 
SF-UDA is highly relevant for applications, since it allows the design of efficient algorithms while, as we show in our experiments, achieving comparable performance to UDA. 
In our analysis, we decouple the effects on the final result of the two adaptations into play in  double-transfer: from pre-training to source domain and from source to target domain. 
With our results, we provide best practices for robust SF-UDA pipelines. Our empirical analysis includes more than 500 architectures (including both CNNs and Vision Transformers). We test them on 6 datasets for domain adaptation (for a total of 74 domain shifts). To the best of our knowledge, this is the most extensive study on SF-UDA.

\section{Methods}

\subsection{Probing}
For our initial experiments we adopt two probing methods to evaluate the quality of features extracted from the models: linear probing and cluster probing.\\
\textbf{Linear Probing (LP)}. 
The pre-trained feature extractor is fixed and employed to compute features from training images. 
These are employed to train a linear classifier (i.e., a multinomial regressor). Finally, classification accuracy is evaluated on the test set. This method is commonly used in the literature to evaluate feature extractors pre-trained with self-supervision~\citep{chen2021empirical}.\\
\textbf{Cluster Probing (CP)}. 
As in LP, we fix the feature extractor and compute features from the training images.
Then, for each class, the \textit{class prototype} is computed as the average feature vector of the examples from that class. 
At prediction time, new samples are classified based on the class of the closest prototype (in our case, we use the cosine dissimilarity). 
As in LP, we evaluate generalization capabilities on the test set. 
CP allows inspecting the properties of the learned representation, 
providing useful insights for SF-UDA which exploits the 
underlying structure of the feature space
to overcome the lack of ground truth on the target domain. Refer to App.~\ref{prob_methods} for more details.\\
\textbf{Remark.} In next sections, we append  ``Gen" or ``DGen" to the method name to indicate the test accuracy on the source or target domain, respectively. If nothing is indicated we use LP as default.
\subsection{SF-UDA}
\label{sf_uda}

\textbf{Simple Class Alignment (SCA).} This algorithm adapts the classifier to the target domain without altering the feature extractor. 
As in \textit{cluster probing}, a first phase computes class \textit{prototypes} of labeled data of the source domain. In the second step, the prototypes are used as initialization centroids in spherical k-means~\citep{hornik2012spherical}, which is executed on the target unlabeled data. 
Hence, the final centroids are adapted prototypes that account for the domain shift.
The resulting classifier assigns 
the class of the closest centroid to new inputs (based on cosine dissimilarity). 
This method was presented by \cite{kang2019contrastive} to compute pseudo-labels. 
Variants are discussed in App.~\ref{scaappendix}.\\
\textbf{Source HypOthesis Transfer (SHOT).} This algorithm~\citep{liang2020we, liang2021source} is considered the state-of-the-art in SF-UDA for its efficiency and it is a solid baseline for novel SF-UDA methods. 
The first transfer requires fine-tuning the model to the source domain. 
The second transfer alternates two steps iteratively: (1) pseudo-labels computation for target samples, and (2) feature extractor fine-tuning using the Information Maximization loss on previously computed pseudo-labels, while keeping the classifier fixed. 
Interestingly, the latter 
proved crucial
for achieving high performance.
For pseudo-labels computation, SHOT builds on a modified version of SCA. 
The feature extractor ($f_\theta$) and the classifier ($h_\phi$), trained on the source domain, are used to predict probabilities for each target sample $x_i$ for all the $C$ classes $p_i = (p_{i1}, \ldots, p_{iC})$.
Then, the initialization prototype for class $c, c=1, \dots, C$ is computed as a weighted average of the target features:

\vspace{-0.3cm}
\begin{equation}
    k_c = \frac{\sum_{x_i \in S_{tgt}} p_{ic} f_{\theta}(x_i)}{\sum_{x_i \in S_{tgt}} p_{ic}}.
\end{equation}
The prototypes $k_1, k_2, \ldots, k_C$ are used to initialize spherical k-means (as in SCA), and the final 1-NN classifier  computes the pseudo-labels of the target dataset.\\
\textbf{Remark.} We stress that an overview and comparison of different SF-UDA approaches is out of the scope of this work. We select SCA and SHOT for our experiments since they are recent representative SF-UDA approaches that tackle the second transfer in two complementary ways. 
Indeed, while SHOT keeps the classifier fixed and adapts the feature extractor, SCA does exactly the opposite. Further experiments with the NRC method~\citep{yang2021exploiting} corroborate our results.

\section{Experiments}

\subsection{Setup}
\textbf{Models.} We rely on the \textit{PyTorch Image Models} Python library (timm)~\citep{rw2019timm} for the implementation of the various models considered in our experiments.
It provides access to a remarkably large number of different architectures and pre-trained weights. 
In our evaluation of the probing approaches and SCA (without fine-tuning) we used 500 models taken from more than 25 different families of architectures (e.g., VGG~\citep{vgg}, ResNet~\citep{resnet}, EfficientNet~\citep{efficientnet}, ConvNext~\citep{convnext}, ViT~\citep{dosovitskiy2020image}, SWIN~\citep{swin}, Deit~\citep{deit} and XCiT~\citep{xcit}). Instead, for the experiments with SHOT and with networks fine-tuning, we sampled a subset of 59 models, taken from more than 12 families of architectures. Notably, our analysis comprises both modern Vision Transfomers and more traditional CNNs. 
More details are available in App.~\ref{models}.\\
\textbf{Pre-training.} For the first transfer, we consider two datasets: ImageNet (ILSVRC-2012), composed of 1.2M images for 1000 mutually exclusive classes, and the superset ImageNet21k~\citep{deng2009imagenet}, composed of 14M images for 21,841 not mutually exclusive classes. 
Specifically, we either consider models pre-trained on ImageNet (\textit{IN}) or on ImageNet21k and then fine-tuned on the 1000 classes of ImageNet (\textit{IN21k}).\\
\textbf{Domain Adaptation Image Datasets.} DA datasets for image classification typically have two or more sub-datasets corresponding to different domains sharing the same classes.
 In our experiments, we considered:
 \textit{DomainNet}~\citep{peng2019moment}, a large dataset with 6 domains of common objects, divided into 345 categories ($596\,006$ images);
 \textit{ImageClef-DA}~\citep{long2017deep}, a small dataset with 4 domains and 12 classes ($2\,400$ images);
 \textit{Modern Office-31}~\citep{ringwald2021adaptiope}, a novel version of the Office-31 dataset~\citep{saenko2010adapting} with an additional synthetic domain (for a total of 4) and 31 classes ($7\,210$ images);
 \textit{Visda-2017}~\citep{peng2017visda}, a large sim-to-real dataset  with 12 categories and 2 domains ($280\,000$ images);
 \textit{Office-Home}~\citep{venkateswara2017deep}, with 4 domains and 65 categories ($15\,588$ images);
 \textit{Adaptiope}~\citep{ringwald2021adaptiope}, having 3 domains (\textit{synthetic}, \textit{product} and \textit{life}) and 123 different classes ($36\,900$ images).
 We end up with 23 different domains and 74 domain pairs in total\footnote{In Visda-2017, differently from the common benchmark, we consider both experiments (synthetic $\to$ real and real $\to$ synthetic).}.\\
\textbf{Experimental Details.} 
We use the aforementioned datasets to conduct a systematic study on the main design choices in  \textit{double-transfer}. Specifically, to evaluate the generalization capability (Gen) of the various feature extractors (first transfer), we randomly split the images from each domain into a training (80\%) and a test set (20\%). 
We use the former to either train the classifier (LP or CP) or to fine-tune the model and we evaluate the accuracy on the test set. The final performance of one feature extractor is given by the average of all the accuracy values obtained on the 23 domains.
Similarly, for the second transfer (DGen and SF-UDA settings), we consider each domain pair and the final performance is obtained by averaging over all the 74 pairs.
Note that for SF-UDA we consider different combinations for the two transfers. 
For the first one, we either train a classifier on the source domain, keeping the feature extractor fixed (as in LP and CP), or we fine-tune (FT) it while training the classifier.
Instead, for the second phase we mainly consider SHOT and SCA. 
This leads to 4 different combinations: SCA, FT+SCA, SHOT and FT+SHOT. 
Finally, in SF-UDA, as it is common in the literature and in UDA benchmarks, we consider the transductive setting~\citep{kouw2019review} where the accuracy is evaluated on the same images used for  adaptation (albeit without labels). 
For technical details on the experiments refer to App.~\ref{techdet}.
\subsection{Relevance of Pre-training and ImageNet Accuracy in SF-UDA}
\label{modselection}
\begin{figure}[t]
\begin{center}
\includegraphics[width=0.85\textwidth]{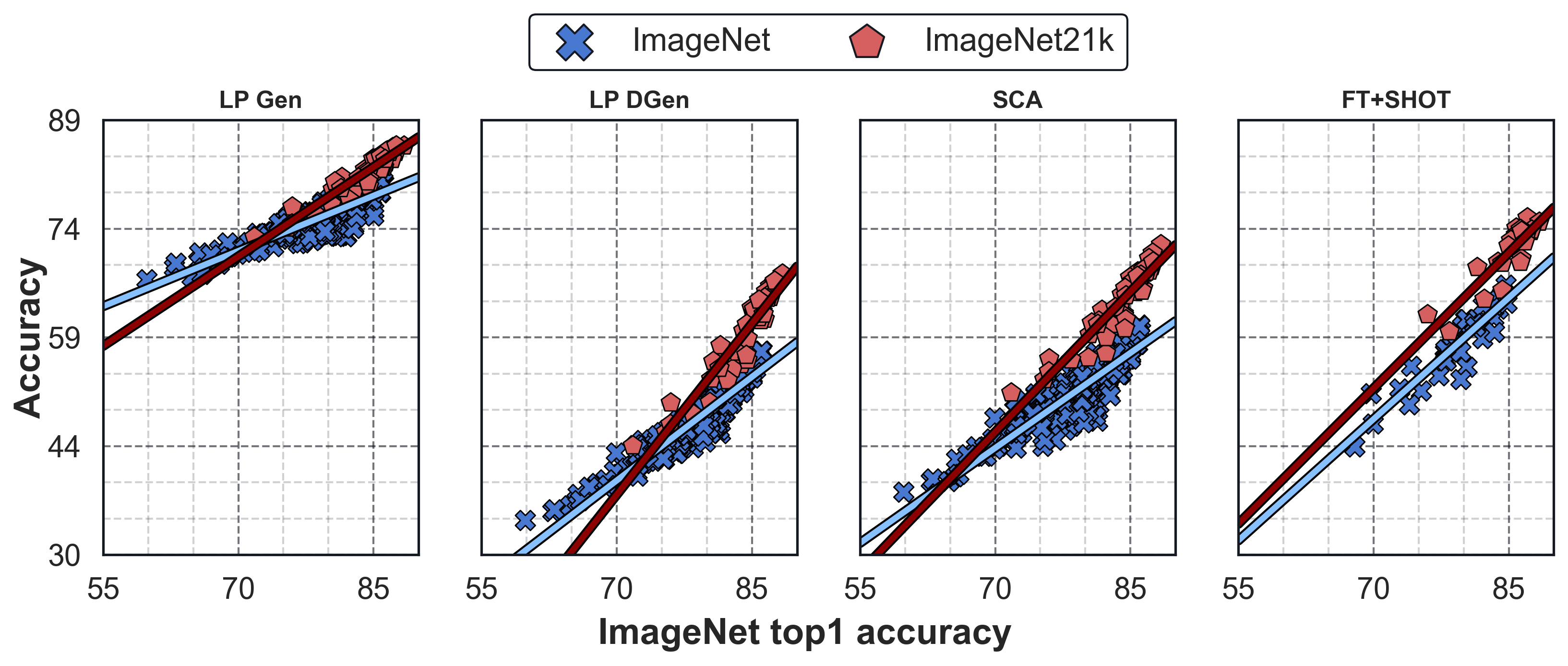}
\end{center}
\caption{We compare performance of different feature extractors (which have been pre-trained either on ImageNet or Imagenet21k) for, respectively: (i) Generalization LP, (ii) Domain generalization LP, (iii) SCA and (iv) FT + SHOT.
In the sub-plot on the left (LP Gen) the test accuracy on the source is considered, while on other subplots (LP DGen, SCA and FT+SHOT) the test is performed on the target domain.
}
\label{multiplot}
\vspace{-0.3cm}
\end{figure}

\begin{table}[t]
\caption{Adjusted $\bar{R}^2$ values for linear (only top1 accuracy considered) and multi-linear (top1 and pre-training) explanatory models for linear probing (LP), cluster probing (CP), SHOT, SCA and fine tuning (FT).} 
\label{table_R}
\begin{center}
\begin{adjustbox}{max width=0.65\textwidth}
\begin{tabular}{lccccc}

\toprule
& \textbf{LP Gen} & \textbf{CP Gen} & \textbf{LP DGen} & \textbf{CP DGen} & \textbf{SCA}\\
\midrule
\textbf{Linear} & 0.736 & 0.756 & 0.810 & 0.755 & 0.731 \\
\textbf{Multi-Linear} & 0.851 & 0.908 & 0.935 & 0.918 & 0.902 \\
\midrule
\midrule
& \textbf{SHOT} & \textbf{FT Gen} & \textbf{FT DGen} & \textbf{FT+ SCA} & \textbf{FT+ SHOT}\\
\midrule
\textbf{Linear} & 0.792 & 0.803 & 0.698 & 0.668 & 0.838\\
\textbf{Multi-Linear} & 0.890 & 0.878 & 0.822 & 0.792 & 0.932\\
\bottomrule

\end{tabular}
\end{adjustbox}
\end{center}
\vspace{-0.4cm}
\end{table}

\textit{How to choose the best feature extractor for SF-UDA?}

In Fig.~\ref{multiplot}, we compare the LP performance of different feature extractors for Generalization and Domain Generalization.
Furthermore, we present results for SCA and FT+SHOT. 
Both ImageNet and ImageNet21k pre-trainings are considered. 
We plot the relationship between the accuracy obtained on ImageNet and the one achieved on the task of interest. 
As it can be noticed, the latter is mainly influenced by two factors: (i) the ImageNet top-1 accuracy, and (ii) the pre-training dataset.
Specifically, in all cases the performance on the task of interest depends linearly on the ImageNet top-1 accuracy of the backbone. 
For instance, the accuracy of a backbone $\mathcal{B}$ can be modeled as:
\begin{equation}
\label{linear_model}
    accuracy(\mathcal{B}) = m \cdot top1(\mathcal{B}) + q + \epsilon,
\end{equation}
\noindent where $m$ and $q$ are the parameters of the linear model (specific to each experiment), while $\epsilon$ is a random variable that accounts for the variance in the data, not explained by the model.\\
Instead, regarding the dataset choice for pre-training, it is known from the literature~\citep{dosovitskiy2020image} that using ImageNet21k can boost the accuracy on ImageNet. 
Thus, according to Equation~\ref{linear_model} this should also improve the accuracy for all the downstream tasks. 
This is confirmed by our experiments. Notably, by taking two models with the same ImageNet top-1 accuracy but pre-trained on different datasets, it is evident from Fig.~\ref{multiplot} that the model pre-trained on ImageNet21k has better performance (on average). 
This implies that the ImageNet21k pre-training yields an additional improvement in the considered transfer tasks that cannot be explained by the increased ImageNet accuracy of the model.
To account for this, we introduce the pre-training into the linear statistical model and describe the data through a multi-linear model with interaction. 
Let $pretrain$ be equal to $1$ for ImageNet21k backbones and to $0$ for ImageNet backbones, then the model becomes: 
\begin{equation}
    accuracy(\mathcal{B}) = [m + \Delta m \cdot pretrain(\mathcal{B})] \cdot top1(\mathcal{B}) + q + \Delta q \cdot pretrain(\mathcal{B}) + \epsilon,
\end{equation}

where $\Delta m$ and $\Delta q$ are the newly introduced parameters in the model.\\
In Tab.~\ref{table_R}, we compare the goodness-of-fit (adjusted $\bar{R}^2$) of the multi-linear model with the linear one.
The difference is significant for all experiments with probing  and SF-UDA methods.
App.~\ref{statanalysis} reports the coefficients of the statistical models and the scatter plots of further experiments.
In addition, in App.~\ref{semantictrain} we compare the performance of the same  architecture (ResNet50) pre-trained on ImageNet21k in 2 ways: (i) with the standard training process (used for all experiments in this work), and (ii) with the technique proposed by~\cite{ridnik2021imagenet}, which employs a specific semantic pre-training on a filtered version of ImageNet21k to reduce the strong class imbalance and to tackle the multi-label nature of the dataset. 
We observe that semantic pre-training significantly improves domain generalization, while when applying SF-UDA methods the gap is recovered and the model performs equally well with both pre-trainings. In App.~\ref{sec:ssl}, we 
compare supervised and self-supervised pre-training  strategies.
Despite the promising results, the former still outperform the second.

\textit{Does the size of the model count?} One may expect that only large models (in terms of number of parameters) could benefit from such a large pre-training dataset as ImageNet21k. 
However, in Tab.~\ref{in_vs_21k} we show that this also applies to smaller models.
\smallskip

\begin{table}[t]
\caption{ImageNet vs ImageNet21k pre-training. The accuracy reported are the average of 74 domain shifts.} 
\label{in_vs_21k}
\begin{center}
\footnotesize
\begin{tabular}{r|cc|cc|cc}

\toprule
\textbf{Backbone} (\textbf{params}) & \multicolumn{2}{c|}{\textbf{LP DGen}} & \multicolumn{2}{c|}{\textbf{SCA}} & \multicolumn{2}{c}{\textbf{FT+SHOT}} \\
& \textbf{IN} & \textbf{IN21k}& \textbf{IN} & \textbf{IN21k}& \textbf{IN} & \textbf{IN21k} \\
\textbf{VGG19 (143.7M)}  & 45.2 & \textbf{47.6} & 49.4 & 53.1 & \textbf{56.0} & 55.7\\
\textbf{ResNet50 (25.6M)} & 47.2 & \textbf{51.3} & 49.8 & \textbf{58.2} & 55.9 & \textbf{62.0} \\
\textbf{W-ResNet50 (68.9M)} & 50.5 & \textbf{52.5} & 53.3 & \textbf{58.2} & 62.1 & \textbf{64.0} \\
\textbf{DenseNet161 (28.7M)} & 48.0 & \textbf{52.2} & 52.7 & \textbf{58.2} & 61.6 & \textbf{65.3} \\
\textbf{ConvNext B (88.6M)} & 54.4 & \textbf{65.2} & 58.4 & \textbf{68.5} & 65.1 & \textbf{72.7} \\
\bottomrule

\end{tabular}
\end{center}

\vspace{-0.7cm}
\end{table}
\subsection{Fine-tuning on the Source Domain}
\label{ft_sec}

\begin{figure}[h!]
\begin{center}
\includegraphics[width=0.9\textwidth]{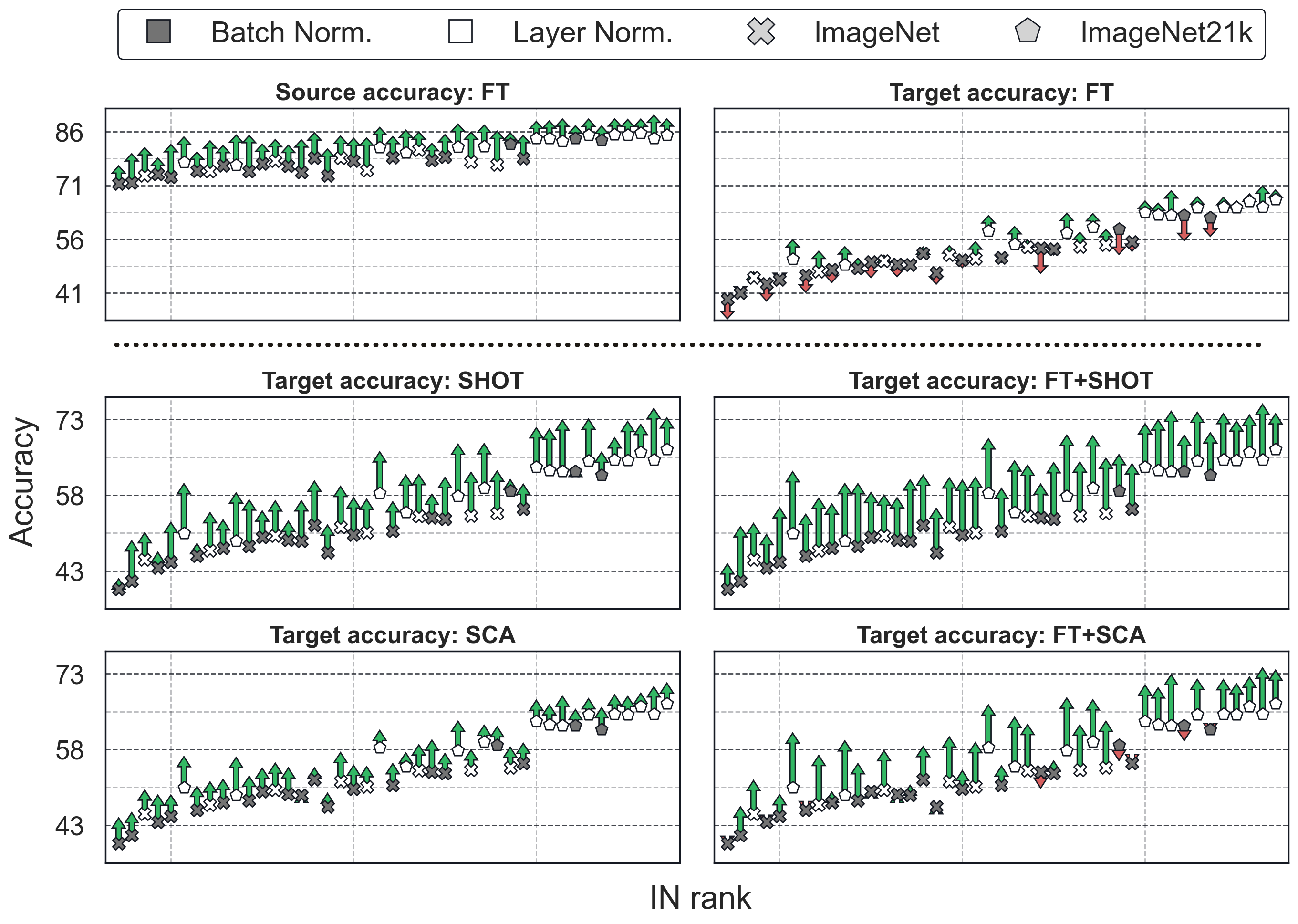}
\end{center}
\caption{Architectures are sorted on the x-axis according to IN top1 accuracy (best ones on the right). We distinguish models based on the employed pre-training and normalization layers. The markers' y-value
represents the LP accuracy of the model, while the green/red arrows represent the accuracy improvement/degradation  when a method is applied. 
FT and LP are always performed on the source with labels, while adaptation methods are always performed on the target without labels.} \textbf{Top-left}: 
source accuracy
improves with fine-tuning (average over 23 domains). \textbf{Top-right}:
target accuracy slightly increases for layer-norm models but it is degraded for batch-norm models. \textbf{Bottom}: SCA and SHOT improve after fine-tuning. In both cases we average results over 74 domain pairs (we report a subset of the models for visualization purposes).
\label{bars_ft}
\vspace{-0.3cm}
\end{figure}

In this section, we study the impact on downstream tasks of fine-tuning the model on the source data (first transfer). 
Results for 59 models are reported in Fig.~\ref{bars_ft} averaged over 23 domains in case of experiments with 
Gen tasks and on the 74 domain pairs in all other cases. 
The reported plots present accuracy differences between each method and its baseline (as green or red arrows). 
For instance, 
the top-left plot
reports the difference in source accuracy between no fine-tuning (i.e., LP) and fine-tuning on the source domain.

Firstly, it can be noticed that in cases where no fine-tuning is applied (namely SHOT and SCA plots), a remarkable accuracy gain is still obtained with respect to LP DGen. Interestingly, Tab.~\ref{avg_gain} highlights how SHOT (which adapts the feature extractor on the target domain) obtains an accuracy gain of $6.97 \% (\pm 2.24)$ on average on the target domain. In comparison, SCA gains on average $4.21 \% (\pm 1.39)$, while being 50-60 times faster than SHOT (see App.~\ref{comptimes}).

Then, we consider the case where fine-tuning is applied on the source domain. It is reasonable to expect that, in the Gen task case, fine-tuning the feature extractor on one domain increases performance on the same domain~\citep{kornblith2019better}. 
However, this does not always improve performance on the second transfer, as can be noticed 
in the second column of Fig.~\ref{bars_ft}. 
Specifically, while for SHOT fine-tuning during the first transfer always 
yields a
subsequent performance gain, models with BN layers do not benefit from this adaptation on the source for SCA and especially for the FT-only target accuracy. 
In some cases, fine-tuning on the source even degrades performance (see App.~\ref{sec:additional_results} for further results).

\begin{figure}[h!]
\begin{center}
\includegraphics[width=0.9\textwidth]{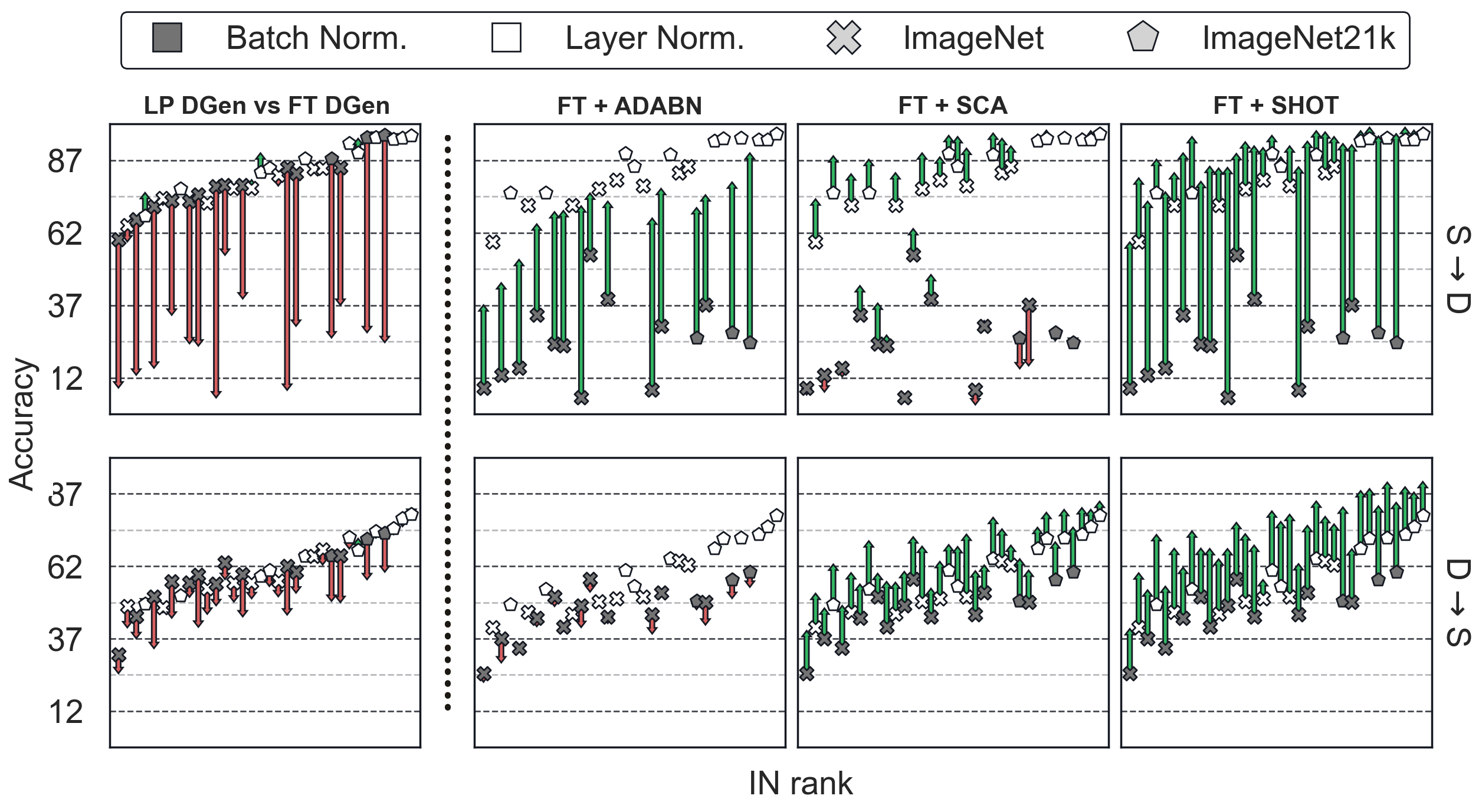}
\end{center}
\caption{Rows represent different domain pairs (Sythetic $\to$ DSLR and DSLR $\to$ Synthetic from Modern Office31). 
In column 1, arrows indicate the effect of FT on the target accuracy with respect to the LP DGen baseline (the markers). In the other columns, the markers represent the target domain accuracy after fine-tuning and arrows represent the performance difference using ADABN, SCA or SHOT (after the fine-tuning).
}
\label{failures}
\end{figure}

We 
thoroughly
analyze this matter in Fig.~\ref{failures}.
We report results for domain pairs from  ModernOffice31 (namely, Synthetic and DSLR), for which this phenomenon is most evident. 
However, it occurs also in a large number of other domain pairs.
In 
the first row, we experiment in the  Synthetic~$\to$~DSLR case. 
As evidenced by the first-column plots, BN layers in this case lead to 
a very significant target accuracy degradation after fine-tuning compared to na\"ive LP. Nonetheless, in the second column we show that using 
ADABN~\citep{li2016revisiting} 
to adapt the BN layers on the target domain 
compensates for the initial loss.
Concerning SF-UDA, note that while SCA is strongly affected by the performance degradation brought by BN layers adaptation (third column), SHOT is able to 
nicely
recover (fourth column), since it employs BN statistics computed on the target domain during the iterative
feature extractor adaptation. 
Finally, in the second row we report results for the  DSLR $\to$ Synthetic transfer. 
Notably, in this case ADABN  does not help recover performance, while both SCA and SHOT allow to compensate for the initial loss. 
Note that there may exist methods to improve the target accuracy after fine-tuning of models with BN (e.g., fixing BN statistics, using augmentations, etc.), but the study of these techniques is out of the scope of our analysis. 

\textbf{Remark 1.} One could ascribe the performance drop 
to architectural properties other than the BN layers, but with the previous examples (i.e., column 2 in Fig.~\ref{failures}) we show that the main cause actually lies in the BN statistics. 
Moreover, the issue never occurs for models with LN layers.\\
\textbf{Remark 2.} The performance drop when fine-tuning models with BN layers is especially evident for pairs with shifts between real and synthetic domains,  but it is not limited to such cases (see, e.g., Clipart $\to$ Art (OfficeHome) and Webcam $\to$ Amazon (ModernOffice31)). 
In App.~\ref{app_ln_vs_bn} we show a more extensive comparison between models with BN and LN layers.\\
\textbf{Remark 3.} Tab.~\ref{avg_gain} reports the average difference in accuracy  with respect to LP DGen, grouping the models used for this experiment by pre-training dataset and normalization layers. 
We compare performance for FT DGen, SCA, FT + SCA, SHOT, and FT + SHOT.

\begin{table*}[t]
\caption{Average accuracy change (\%) with respect to DGen with LP. For each architecture, the accuracy change is the average over 74 different domain-shifts.
Each column shows the performance obtained by models grouped by different combinations of their pre-training and the normalization layers. The number of considered models is reported in parenthesis. In each row, we report results for different downstream tasks.}
\label{avg_gain}
\begin{center}
\scriptsize
\setlength\tabcolsep{3.5pt}
\begin{adjustbox}{max width=0.99\textwidth}
\begin{tabular}{l|c|cc|cc|cc|cc}
\toprule
\diagbox{\textbf{Tasks}}{\textbf{Models}} & \textbf{All} & \textbf{IN} (32) & \textbf{IN21k} (27)  & \textbf{BN} (32) & \textbf{LN} (27) & \textbf{IN+BN} (24) & \textbf{IN+LN} (8) & \textbf{IN21k+BN} (8) & \textbf{IN21k+LN} (19) \\
\midrule
\textbf{FT DGen} & 0.46 \tiny{$\pm$ 3.71} & -0.39 \tiny{$\pm$ 3.23} & 1.46 \tiny{$\pm$ 4.04} & -2.21 \tiny{$\pm$ 2.75} & 3.62 \tiny{$\pm$ 1.57} & -1.61 \tiny{$\pm$ 2.66} & 3.28 \tiny{$\pm$ 1.55} & -4.02 \tiny{$\pm$ 2.30} & 3.77 \tiny{$\pm$ 1.59}\\
\textbf{SCA} & 4.21 \tiny{$\pm$ 1.39} & 3.97 \tiny{$\pm$ 1.40} & 4.49 \tiny{$\pm$ 1.35} & 4.08 \tiny{$\pm$ 1.55} & 4.36 \tiny{$\pm$ 1.19} & 3.75  \tiny{$\pm$ 1.52} & 4.64 \tiny{$\pm$ 0.66} & 5.07  \tiny{$\pm$ 1.27} & 4.25 \tiny{$\pm$ 1.35}\\
\textbf{SHOT} & 6.97 \tiny{$\pm$ 2.24} & 6.57 \tiny{$\pm$ 1.95} & 7.45 \tiny{$\pm$ 2.51} & 6.21\tiny{$\pm$ 2.39} & 7.88 \tiny{$\pm$ 1.68} & 6.29  \tiny{$\pm$ 2.10} & 7.42 \tiny{$\pm$ 1.12} & 5.97  \tiny{$\pm$ 3.29} & 8.07 \tiny{$\pm$ 1.86}\\
\textbf{FT+SCA} & 4.90 \tiny{$\pm$ 3.86} & 4.20  \tiny{$\pm$ 3.62} & 5.72 \tiny{$\pm$ 4.03} & 2.16 \tiny{$\pm$ 2.94} & 8.14 \tiny{$\pm$ 1.66} & 2.74  \tiny{$\pm$ 2.91} & 8.59 \tiny{$\pm$ 0.98} & 0.44  \tiny{$\pm$ 2.45} & 7.95 \tiny{$\pm$ 1.86}\\
\textbf{FT+SHOT} & 9.66 \tiny{$\pm$ 1.83} & 9.68 \tiny{$\pm$ 1.76} & 9.63 \tiny{$\pm$ 1.93} & 9.68 \tiny{$\pm$ 1.96} & 9.63 \tiny{$\pm$ 1.69} & 9.67  \tiny{$\pm$ 1.90} & 9.71 \tiny{$\pm$ 1.40} & 9.72  \tiny{$\pm$ 2.28} & 9.60 \tiny{$\pm$ 1.84}\\
\bottomrule
\end{tabular}
\end{adjustbox}
\end{center}
\vspace{-0.5cm}
\end{table*}
\smallskip

\subsection{Robustness and Failures Analysis }
\label{robust_sec}

\begin{table*}[t]
\caption{Average Failure Rate (\%) on 74 domain shifts. Each column shows the performance obtained by models grouped by different combinations of their pre-training and the normalization layers. The number of considered models is reported in parenthesis. In each row, we report results for different downstream tasks.} 
\label{failure_rates}
\begin{center}
\scriptsize
\setlength\tabcolsep{3.5pt}
\begin{adjustbox}{max width=0.99\textwidth}
\begin{tabular}{l|c|cc|cc|cc|cc}
\toprule
\diagbox{\textbf{Tasks}}{\textbf{Models}} & \textbf{All} & \textbf{IN} (32) & \textbf{IN21k} (27)  & \textbf{BN} (32) & \textbf{LN} (27) & \textbf{IN+BN} (24) & \textbf{IN+LN} (8) & \textbf{IN21k+BN} (8) & \textbf{IN21k+LN} (19) \\
\midrule

\textbf{FT DGen} & 40.82 \tiny{$\pm$ 16.14} & 45.86 \tiny{$\pm$ 13.19} & 34.83 \tiny{$\pm$ 17.48} & 51.52 \tiny{$\pm$ 11.42} & 28.13 \tiny{$\pm$ 10.85} & 50.51 \tiny{$\pm$ 11.26} & 31.93 \tiny{$\pm$ 7.61} & 54.56 \tiny{$\pm$ 12.11} & 26.53 \tiny{$\pm$ 11.77}\\

\textbf{SCA} & 21.05 \tiny{$\pm$ 11.92} & 25.84 \tiny{$\pm$ 12.43} & 15.37 \tiny{$\pm$ 8.42 } & 27.24 \tiny{$\pm$ 11.64} & 13.71 \tiny{$\pm$ 7.25 } & 28.89 \tiny{$\pm$ 12.78} & 16.72 \tiny{$\pm$ 4.62} & 22.30 \tiny{$\pm$ 5.11 } & 12.45 \tiny{$\pm$ 7.86 }\\

\textbf{SHOT} & 11.18 \tiny{$\pm$ 8.75 } & 13.13 \tiny{$\pm$ 8.97 } & 8.86  \tiny{$\pm$ 8.03 } & 14.82 \tiny{$\pm$ 9.32 } & 6.86  \tiny{$\pm$ 5.61 } & 14.13 \tiny{$\pm$ 9.11 } & 10.14 \tiny{$\pm$ 8.36} & 16.89 \tiny{$\pm$ 10.27} & 5.48  \tiny{$\pm$ 3.39 } \\

\textbf{FT+SCA} & 19.08 \tiny{$\pm$ 14.57} & 23.35 \tiny{$\pm$ 13.89} & 14.01 \tiny{$\pm$ 13.93} & 29.77 \tiny{$\pm$ 11.05} & 6.41  \tiny{$\pm$ 4.52 } & 29.00 \tiny{$\pm$ 11.19} & 6.42  \tiny{$\pm$ 2.58} & 32.09 \tiny{$\pm$ 11.00} & 6.40  \tiny{$\pm$ 5.19 } \\

\textbf{FT+SHOT} & 4.47  \tiny{$\pm$ 3.61 } & 5.15  \tiny{$\pm$ 3.62 } & 3.65  \tiny{$\pm$ 3.49 } & 5.83  \tiny{$\pm$ 3.77 } & 2.85  \tiny{$\pm$ 2.69 } & 5.52  \tiny{$\pm$ 3.74 } & 4.05  \tiny{$\pm$ 3.23} & 6.76  \tiny{$\pm$ 3.96 } & 2.35  \tiny{$\pm$ 2.33 } \\
\bottomrule

\end{tabular}
\end{adjustbox}
\end{center}
\end{table*}

In SF-UDA, target annotations are unavailable, precluding performance evaluation after adaptation in real scenarios.
Therefore, it would be beneficial in practice to have an estimate of how likely SF-UDA methods are to degrade performance on the target domain 
and
how severe  such degradation could be.
To this aim, we 
extensively evaluate
the robustness of SF-UDA.
Given a model and a DA task, we fix 
LP DGen
as baseline
 and consider an SF-UDA approach to fail
if its 
target accuracy 
after adaptation
is lower than 
the baseline's. 
Tab.~\ref{failure_rates} groups 59 models 
by
pre-training dataset and normalization layers and reports the average failure rates 
of each group on the 74 considered domain pairs.
BN models always
show a remarkably higher 
failure rate: 
FT 
degrades the target accuracy 51.5\% of the times for BN models, while just 28.1\% for LN ones. 
Applying SCA or SHOT after fine-tuning can reduce failures, but  LN models are still more robust and fail less frequently.
Pre-training also impacts the failure rate: LN models pre-trained on ImageNet21k have a lower rate than those pre-trained on ImageNet.
However, BN models deteriorate more often when pre-trained on ImageNet21k. 
Indeed,
ImageNet21k models, as 
shown in Sec.~\ref{modselection}, achieve a better LP DGen, and  therefore a stronger baseline value.
For this reason, the instability of BN layers might result in a more apparent degradation. 
Finally, note that SCA is very sensitive to the normalization layers choice and that FT+SHOT is the method with the lowest  failure rate overall.\\
\textbf{Performance Degradation in Case of Failure.} As before, degradation magnitude mostly depends on normalization: in case of failure, performance on target domain after FT (on source) decreases by 10.5\% for BN and by just 2.10\% for LN. SCA is unable to recover the  FT degradation:  in fact, the accuracy of FT+SCA decreases by 12.7\% for BN and only  1.4\% for LN models.
The full table with average performance degradation is reported in App.~\ref{app_ln_vs_bn}.

\subsection{Comparison with Standard UDA \label{sec:vsuda}}
\begin{table}
\centering
\caption{Comparison, using modern architectures (SWIN and ConvNext) of SF-UDA (SCA, SHOT and NRC) with not Source-Free methods: DANN~\citep{ganin2016domain}, CDAN~\citep{cdan}, AFN~\citep{afn}, MDD~\citep{mdd} and MCC~\citep{mcc}.}
\label{table_uda}
\begin{threeparttable}
    \footnotesize
    
    \begin{tabular}{l|cc|cc}
    \textbf{Backbone} & \multicolumn{2}{c|}{SWIN L} & \multicolumn{2}{c}{ConvNext XL} \\
    \midrule
    \textbf{Dataset} & Office-Home & DomainNet  & Office-Home & DomainNet \\
    \midrule
    \textbf{DANN}\tnote{\textdagger} & 86.6 & 49.4 & 86.7 & 48.8 \\ \textbf{CDAN}\tnote{\textdagger} & 88.4 & 50.4 & 89.5 & 51.2 \\ \textbf{AFN}\tnote{\textdagger}  & 85.7 & 46,4 & 85.5 & 46.7 \\ \textbf{MDD}\tnote{\textdagger}  & 86.5 & 41.5 & 86.4 & 42.8 \\ \textbf{MCC}\tnote{\textdagger}  & 88.3 & 47.1 & 88.9 & 45.6 \\
    \midrule
    \textbf{FT+SCA}\tnote{*}  & 86.4 & 49.0 & 87.4 & 49.8 \\ 
    \textbf{FT+SHOT}\tnote{*} & 89.3 & 51.2 & 88.8 & 51.0 \\ 
    \textbf{FT+NRC}\tnote{*}  & 89.5 & 48.4$^\diamond$ & 89.2 & 47.1$^\diamond$ \\
    \bottomrule
    \end{tabular}
    \begin{tablenotes}
        \scriptsize
    
        \item[\textdagger] Requires source, reproduced by~\cite{kim2022broad}.
        \item[*] Source-Free, reproduced by us.
        \item[$\diamond$] Trained and evaluated on a random 15\% subset of DomainNet (official distributed implementation not available).
    \end{tablenotes}
    
\end{threeparttable}
\vspace{-0.5cm}
\end{table}
Recent results indicate overfitting to the benchmark for UDA approaches~\citep{kim2022broad}: with modern backbones, classic algorithms outperform newer ones on uncommon datasets. 
In constrast, SF-UDA techniques demonstrate a performance gain with respect to the LP DGen baseline.\\
In Tab.~\ref{table_uda}, we report SCA, SHOT, and NRC in a setting different from the standard evaluation protocol and compare them to both established and novel UDA solutions. We consider two modern architectures: SWIN~\citep{swin} and ConvNext~\citep{convnext}.
Three main observations arise: (i) despite the stricter constraints on the learning setting, SF-UDA algorithms are competitive with standard UDA; (ii) SCA performs on par (or even outperforms) common UDA algorithms while being
more efficient
(see Tab.~\ref{times} in App.~\ref{comptimes}) and, (iii) although FT+NRC has been trained on a subset of DomainNet, 
this proved sufficient for it to be competitive with UDA counterparts.

\section{Conclusions}
The lack of supervision in UDA hinders model selection when
designing and deploying learning systems.
In this challenging setting, large-scale experiments can provide empirical insights into best practices to tackle real-world UDA problems.
In this work, we shed light on the role of the key design choices underlying SF-UDA algorithms: pre-training dataset, architecture, fine-tuning, and normalization strategy. 
To this end, we present extensive experiments evaluating the impact of such choices.
Our results consistently demonstrate that pre-training accuracy on the standard ImageNet strongly correlates with domain adaptation performance. 
Moreover, despite similar performance on the standard ImageNet test set, models pre-trained on ImageNet21k adapt more effectively. 
Our results challenge the common practice of fine-tuning to the task at hand:
blind adaptation might lead to catastrophic failures, mainly when BN is applied and normalization statistics of the source data are distant from the target ones.
Our robustness analysis confirms the critical role of the normalization layers. 
On average, BN architectures suffer from a higher failure rate, i.e.,  adaptation deteriorates performances. 
Despite being more constrained in terms of source data availability and computational cost, SF-UDA proves competitive with UDA approaches.
Future work will cover tasks beyond image classification and settings with limited access to the source model data, training strategies, and parameters.
\clearpage

\nocite{langley00}
\bibliographystyle{unsrtnat}
\bibliography{ref}

\newpage
\appendix
\onecolumn

\clearpage
\section{Probing methods}
\label{prob_methods}
In this section, we discuss the probing methods employed to inspect and assess learnt models.

We follow the the same notation introduced in Sec.~\ref{related_sec}. Let $f_{\bm{\theta}}$ be a given fixed feature extractor and $C$ the number of classes. We introduce the training set $\mathcal{S}_{\text{TRAIN}} = \{(z_i, y_i)\}_{i=1}^N$, where $z_i = f_{\bm{\theta}}(x_i)$ and $(x_i, y_i) \sim \mu_S$ are \textit{i.i.d.} samples from the source domain, for $i=1, \ldots, N$.

Similarly, we define $\mathcal{S}_{\text{TEST}}$ to be a set of $M$ \textit{i.i.d.} evaluation samples. Based on the experiment, evaluation samples can come from the same distribution of the training (generalization) or from a different one (domain-generalization and domain adaptation).

\textbf{Linear Probing} (LP) trains and evaluates a linear classifier (Multinomial Regression) on the features extracted from a fixed model. During training of the linear classifier, we seek for a function $g$ that maps the features to the $C-1$ simplex minimizing the (L2-regularized) $\log$ loss:

\begin{align}
    g: \mathcal{Z} &\to \Delta^{C-1}, \\
    z &\mapsto \sigma(Wz),
\end{align}

where $\sigma$ is the \textit{softmax} function and $W \in \mathbb{R}^{C \times D}$ is found as:
\begin{equation*}
    \label{findW}
    W = \argmin_{W' \in \mathbb{R}^{C \times D}} \lambda || W' ||_F^2 +  \frac{1}{|\mathcal{S}_\text{TRAIN}|}\sum_{(z, y) \in \mathcal{S}_\text{TRAIN}}-\log(\sigma(W'z)_{y}),
\end{equation*}

with $\lambda > 0$ regularization hyperparameter.\\
Then, we can convert this function to a classifier $h_{MR}$:
\begin{equation*}
\begin{split}
    h_{MR}: \mathcal{Z} &\to \mathcal{Y},\\
    z &\mapsto \argmax_k g(z)_k.
\end{split}
\end{equation*}
Hence, linear probing accuracy is evaluated on the test set:

\begin{equation}
\label{eq:LPaccuracy}
    \text{LP accuracy} = \frac{1}{|\mathcal{S}_\text{TEST}|}\sum_{(z, y) \in \mathcal{S}_\text{TEST}}\mathbbm{1}\{h_{MR}(z) = y\}.
\end{equation}

\textbf{Cluster Probing} (CP): trains and evaluates a 1 Nearest Neighbour (1-NN) classifier on the features extracted from a fixed model. For every class, a prototype is found averaging the feature vectors of that class and, then, the prototypes are used as a 1-NN classifier.
In particular, for each class $c$ the prototype is evaluated as:
\begin{equation*}
    k_c = \frac{\sum_{(z, y) \in \mathcal{S}_\text{train}} \mathbbm{1}\{y=c\}z}{\sum_{(z, y) \in \mathcal{S}_\text{train}} \mathbbm{1}\{y=c\}},
\end{equation*}
where $c=1, \dots, C$. Hence, the associated classifier is:
\begin{equation*}
\begin{split}
\label{one_nn}
h_{1NN}: \mathcal{Z} &\to \mathcal{Y}, \\
z &\mapsto \argmin_{c \in \mathcal{Y}} \left\{\frac{1}{2} - \frac{1}{2}\frac{\langle k_c, z \rangle}{||k_c ||_2 ||z||_2}\right\},
\end{split}
\end{equation*}

where we use the \textit{cosine dissimilarity} to quantify the  distance between features and prototypes. 

The accuracy is evaluated for cluster probing as in equation~\ref{eq:LPaccuracy}, using $h_{1NN}$:
\begin{equation*}
\label{accuracy}
    \text{CP accuracy} = \frac{1}{|\mathcal{S}_\text{TEST}|}\sum_{(z, y) \in \mathcal{S}_\text{TEST}}\mathbbm{1}\{h_{1NN}(z) = y\}.
\end{equation*}

Intuitively, an high accuracy in the cluster probing means that the features of the same class are well clustered (spherically), while clusters of different classes are nicely separated (always spherically). This is a very desirable property for unsupervised domain adaptation where we need to leverage on the underlying structure, since labels of the target domain are not available.

\section{Simple Class Alignment}
\label{scaappendix}
The Simple Class Alignment (SCA) method is part of many domain adaptation algorithms like CAN~\citep{kang2019contrastive} and SHOT~\citep{liang2020we}.
Nevertheless, in literature, the method is never employed alone and its contributions, as part of the overall DA methods, have not been explored extensively.

The algorithm stands out for its simplicity: (1) find a prototype for every class of the source domain, (2) use these prototypes as initialization for K-Means on the unlabeled target domain and finally, (3) use the resulting prototypes as a 1-NN classifier for the target domain, i.e., classify every sample based on the class of the closest prototype. As in previous works, we used cosine dissimilarity (spherical K-Means) to evaluate how close two feature vectors are.

We highlight that this method finds a new classifier for the new domain by aligning the prototypes to the target distribution. Hence, it can be effective with features extracted from a frozen model without fine-tuning. Consequently, SCA is a highly efficient Domain Adaptation method that can achieve results comparable with the state-of-the-art.

But, \textit{how to find the initialization prototypes?} We compare 4 different techniques considering a problem where we have $C$ classes and a fixed feature extractor that can be used to extract, from images, feature vectors in $\mathbb{R}^d$ (we indicate feature vectors with letter $z$).

\textbf{From source labels.} Leverage the available source labels to partition the source feature vectors into the sets $\mathcal{C}_1, \ldots, \mathcal{C}_C$, where all features of class $j$ are in $\mathcal{C}_j$. Then, compute the prototype for class $j$ as the average of features of that class:

\begin{equation}
    w_j = \frac{1}{|\mathcal{C}_j|}\sum_{z \in \mathcal{C}_j} z.
\end{equation}

This initialization has been used in CAN~\citep{kang2019contrastive} method.

\textbf{From Multinomial Regression weights}. Learn a Multinomial Regression (MR) classifier $h: z \mapsto \sigma(Wz)$, where $\sigma$ is the Softmax function, by training on the labeled source domain. Then, initialize the prototypes as the rows of the matrix $W \in \mathbb{R}^{C \times d}$. Algorithm~\ref{alg:cap} shows all the steps of SCA with this initialization.

\textbf{From target hard predictions (pseudo-labels)}. Train a MR classifier on the source domain and pseudo-label the target accordingly. Initialize each prototype with the average feature of samples associated to the same pseudo-label. We note that this initialization, followed by spherical K-Means is very similar to the previous one: the only difference is that, in this case, in the first iteration to compute the distances for the assignments the weights of matrix $W$ and features $z$ are not normalized. As we will see, this method do not perform as good as the others.

\textbf{From target soft predictions}. Fit a MR classifier on source data and compute prototypes as weighted averages of all samples where weights are given by predicted class probabilities. 
More precisely, for each feature target vector $z_i$ the output of the MR classifier (after the Softmax) is $p_i = (p_{i1}, \ldots, p_{iC})$ where $p_{ic}$ is the predicted probability of sample $i$ to belong to class $c$. Then, the prototype of class $c$ is evaluated as:

\begin{equation}
    k_c = \frac{\sum_{i} p_{ic}z_i}{\sum_{i} p_{ic}}.
\end{equation}

This method has been used in SHOT algorithm.

We remark that these methods can also be applied if a fine-tuning of the feature extractor has been performed or a linear layer is employed in place of the MR classifier. 

\begin{algorithm}
\label{clust_alg}
\caption{SCA from Multinomial Regression weights}\label{alg:cap}
\begin{algorithmic}

\Require \\
\begin{itemize}
\item Multinomial regressor for source domain with weights $W = (w_1, w_2, \ldots, w_C)^T \in \mathbb{R}^{C \times d}$.
\item $Z = (z_1, z_2, \ldots, z_N)^T \in \mathbb{R}^{N \times d}$: unlabeled target samples (features).
\end{itemize}
\\
\State $\mathbf{1}_d \gets (1, ..., 1)^T$
\State $X \gets \text{diag}((Z \circ Z)\mathbf{1}_d)^{-1/2}Z$ \Comment{Normalize features to unit length.}
\\
\While{ not converged}
\\
\State $W \gets \text{diag}((W \circ W)\mathbf{1}_d)^{-1/2}W$ \Comment{Normalize weights to unit length.}
\\

\For{$i$ in $1, \ldots, N$} \Comment{Assign every sample to the most similar weight vector.}
    \State $c_i \gets \argmin_{j \in \{1, \ldots, C \}} \langle Z_{i, :} , W_{j, :} \rangle$
\EndFor
\\

\For{$j$ in $1, \ldots, C$} \Comment{Update the weight vectors.}
    \State $W_{j,:} \gets \frac{\sum_{i=1}^{N} \mathbbm{1}\mathrm{\{c_i = j\}}Z_{i,:}}{\sum_{i=1}^{N} \mathbbm{1}\mathrm{\{c_i = j\}}} $
\EndFor
\\
\EndWhile
\\
\\
\Return{$W$}

\end{algorithmic}
\end{algorithm}

\textit{Which is the best method to initialize prototypes?}
In Tab.~\ref{comparison_sca} we report the average gain on all models of SCA (without fine-tuning), the failure rate, the average gain in case of successful adaptation and the average degradation in case of negative transfer for the 4 types of initialization.  The SCA gain/degradation is considered with respect to naive domain generalization.
The method that performs best is the initialization from the Multinomial Regressor weigths: it is more robust (about $20\%$ failure rate) and the degradation, in case of failure, is modest ($1.8\%$).

The method of hard predictions is the only one that performs poorly, when it succeed it can gain more than other methods ($7.2\%$), but it has a large failure rate and it leads to large performance impairment in case of failure. 

\begin{table}[h]
\caption{Comparison of the four different SCA initializations. The results are the average of 548 architectures and 74 different domain pairs. The $\Delta$ accuracy is the difference between the target accuracy after SCA and the naive domain generalization of MR classifier trained on source. The failure rate is the percentage of experiments where SCA degrades the naive domain generalization. In the last two rows are reported the accuracy gain and degradation considering the successful and failure adaptation, respectively. These result differ from the ones in the main text, since here we consider the full set of 548 architectures.}

\label{comparison_sca}
\begin{center}
\begin{footnotesize}
\begin{tabular}{ccccc}

\toprule
 & \textbf{Source Labels} & \textbf{MR Weights} & \textbf{Hard Pred.} & \textbf{Soft. Pred.} \\
 \midrule
$\bm{\Delta}$ \textbf{Accuracy}& 3.8 $\pm$ \tiny{1.2} & 4.5 $\pm$ \tiny{1.1} & -0.3 $\pm$ \tiny{2.1} & 3.7 $\pm$ \tiny{1.5} \\
\textbf{\textbf{Failure Rate}} & 25.1 $\pm$ \tiny{10.2} & 20.1 $\pm$ \tiny{7.2} & 46.2 $\pm$ \tiny{8.1} & 29.3 $\pm$ \tiny{10.2} \\
 \midrule
$\bm{\Delta}$ \textbf{Accuracy} | \textbf{Success} & 5.8 $\pm$ \tiny{1.0} & 6.2 $\pm$ \tiny{0.9} & 7.2 $\pm$ \tiny{0.1} & 6.4 $\pm$ \tiny{1.0} \\
$\bm{\Delta}$ \textbf{Accuracy} | \textbf{Failure} & -1.8 $\pm$ \tiny{0.6} & -1.8 $\pm$ \tiny{1.0} & -15.4 $\pm$ \tiny{4.3} & -2.4 $\pm$ \tiny{1.6} \\
 \bottomrule
 
\end{tabular}
\end{footnotesize}
\end{center}
\end{table}


\section{Architectures}
\label{models}
\begin{figure}[h!]
\begin{center}
\includegraphics[width=0.95\textwidth]{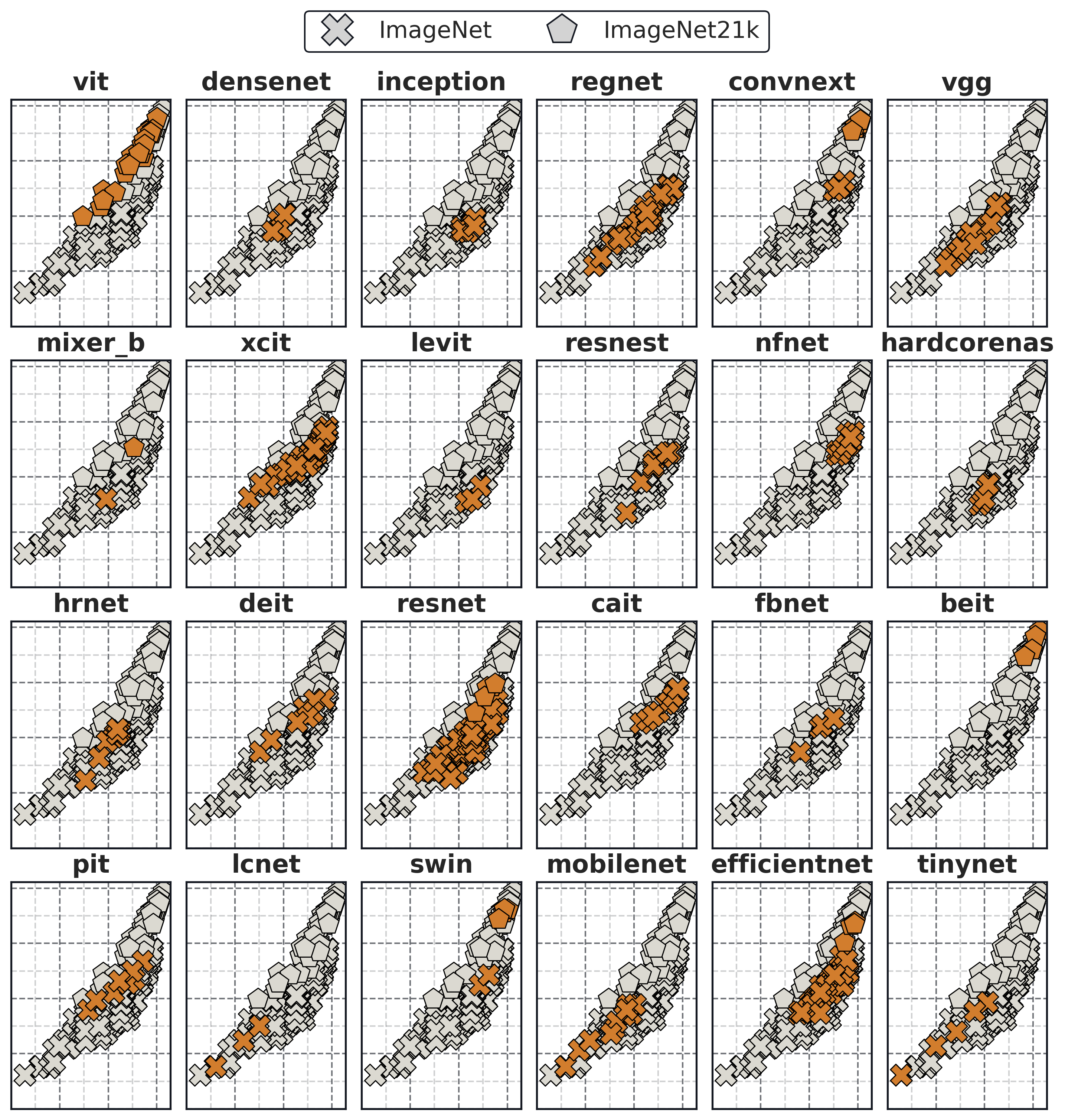}
\end{center}
\caption{Families of architectures: on the x-axis we have the ImageNet top1 accuracy, on the y-axis the accuracy after applying SCA (without fine-tuning) averaged across 74 domain shifts.}
\label{reg1}
\end{figure}

In Fig.~\ref{reg1} we present some plots to help the localization of architecture families on the scatter plots. The architectures of the same family differs for size, e.g., Resnet50 and Resnet101, training procedure (even if the training is always performed on ImageNet or ImageNet21k) or for minor architectural details.

\textbf{Subset of 59 models.} For fine-tuning and SHOT experiments we used a subset of 59 models taken from the  timm library (with the exception of resnet50\_in21k, resnext101\_32x8d\_in21k, densenet161\_in21k, wide\_resnet50\_2\_in21k, vgg19\_bn\_in21k, that were trained by us). 
The models are:

beit\_base\_patch16\_384, beit\_large\_patch16\_224, beit\_large\_patch16\_384, convnext\_base, convnext\_base\_in22ft1k, convnext\_large, convnext\_xlarge\_384\_in22ft1k, convnext\_xlarge\_in22ft1k, deit\_base\_distilled\_patch16\_224, deit\_small\_distilled\_patch16\_224, dla169, ecaresnet101d, efficientnetv2\_rw\_m, efficientnet\_b4, efficientnet\_el, efficientnet\_em, efficientnet\_es, fbnetv3\_d, ghostnet\_100, gluon\_resnet50\_v1c, gluon\_resnext101\_32x4d, gmixer\_24\_224, hrnet\_w44, jx\_nest\_tiny, mixer\_b16\_224\_miil, mixnet\_m, pit\_ti\_224, pit\_xs\_distilled\_224, repvgg\_b3g4, resnet152d, resnet18, resnet26, resnet50, selecsls60, swin\_base\_patch4\_window12\_384, swin\_base\_patch4\_window7\_224, swin\_large\_patch4\_window7\_224, tf\_efficientnetv2\_l\_in21ft1k, tf\_efficientnetv2\_s\_in21ft1k, tf\_efficientnetv2\_xl\_in21ft1k, tf\_mobilenetv3\_small\_100, vgg19\_bn, visformer\_small, vit\_base\_patch16\_224\_miil, vit\_base\_patch8\_224, vit\_base\_r50\_s16\_384, vit\_large\_patch16\_224, vit\_large\_patch16\_384, vit\_large\_patch32\_384, vit\_small\_patch16\_384, vit\_small\_patch32\_224, vit\_tiny\_patch16\_384, xception65, xception71, resnet50\_in21k, resnext101\_32x8d\_in21k, densenet161\_in21k, wide\_resnet50\_2\_in21k, vgg19\_bn\_in21k.

\section{Technical details}
\label{techdet}

For all experiments we used Python 3.7.6 with PyTorch 1.12.1 and CUDA 10.2.

We followed the common procedure used in SF-UDA (see for example SHOT~\cite{liang2020we}) with some adjustments to face the large number of experiments and the great variety and diversity of datasets and models.

\textbf{Model.} For every pre-trained model we remove the last linear layer (ImageNet classifier). Then, we add a randomly initialized bottleneck followed by a linear layer (classifier for the given task).
The bottleneck is composed by a linear layer that maps the features of the backbone to 256 dimensions, followed by a normalization layer and a ReLU activation. The bottleneck normalization layer depends on the backbone, i.e., Batch Normalization if the backbone contains Batch Normalization layers, otherwise it is a Layer Normalization layer.

\textbf{Training.} The distributed training is performed on a variable number of GPUs depending on the model size (between 2 and 16 Nvidia V100 16GB) and automatic mixed precision is used by default.
The batch normalization statistics are always synchronized between different processes.
The global batch size is kept fixed at 64 and, in the cases where the memory of 16 GPUs is not enough, gradient accumulation is used. SGD with Nesterov momentum (0.9) is used for the optimization. The initial learning rate for the classifier and the bottleneck is 0.01, while the initial learning rate of the pre-trained backbone is 0.001. The learning rates are decreased following the same exponential scheduling used in SHOT.
During training, the 15\% of the training set is kept as validation (used just to determine the stopping iteration). To face the great diversity of dataset sizes we limit the number of optimization steps of each epoch to 100 (in practise we considered one epoch just the interval between two validations and not, as it is usual, an iteration of the full dataset). 
We fixed the minimum number of epochs at 10 and the maximum number at 100. After each epoch the validation accuracy is evaluated and if there are 5 epochs in a row where the validation accuracy does not increase, the training is stopped and the model weights with the highest validation accuracy are returned.
As regularization we used weight decay (0.001) and label smoothing (0.1) on the standard crossentropy loss.

\textbf{SHOT.} For the adaptation with SHOT we limited the number of epochs at 15 (always with a maximum of 100 optimization steps) and we used the same hyperparameters of the original paper. We always used distributed training and automatic mixed precision.

\textbf{NRC.} For NRC~\citep{yang2021exploiting} method we used the official code (not distributed) with the official hyperparameters. We added just automatic mixed precision to allow the training of larger models.

\textbf{ImageNet21k training.} To train ResNet50, VGG19, Wide-Resnet50, Resnext101 32x8d and DenseNet161 on ImageNet21k we used the official torchvision training script with the default hyperparamters, with the exception of:
\begin{itemize}
    \item initial learning rate: 0.8
    \item batch size: 2048
    \item lr schedule: constant decay every of 0.1 every 15 epochs
    \item epochs: 50 
\end{itemize}
No augmentations are used and neither a validation set, the final weights are tested on the downstream SF-UDA tasks.
\clearpage

\section{Statistical analysis of results}
\label{statanalysis}
\textit{Are the pre-training differences that we observed relevant?} Even if, from the scatter plots presented in Fig.~\ref{multiplot} it is possible to see, visually, that there is an interesting gap between models pre-trained on ImageNet and ImageNet21k we fit a linear model to understand the benefits of introducing pre-training as independent variable.
In particular, for any experiment, first we try to associate the result accuracy that it is possible to obtain with a given backbone $\mathcal{B}$ to its ImageNet top1 accuracy:
\begin{equation}
    accuracy(\mathcal{B}) = m \cdot top_1(\mathcal{B}) + q
\end{equation}

It is possible to find the parameters $m$ and $q$ with the ordinary least squares estimator. 
Then, we encode the two pre-training into a binary function such that if the backbone is pre-trained on ImageNet $pretrain(\mathcal{B}) = 0$, while if it is pre-trained on ImageNet21k $pretrain(\mathcal{B}) = 1$ and we introduce it into the model. In particular we can consider the following:

\begin{itemize}
    \item \textbf{Different intercept model}
    \begin{equation}
        accuracy(\mathcal{B}) = m \cdot top_1(\mathcal{B}) + (q + \Delta q \cdot pretrain(\mathcal{B}))
    \end{equation}
    
    \item \textbf{Different slope model (interaction)}
    \begin{equation}
         accuracy(\mathcal{B}) = (m + \Delta m \cdot pretrain(\mathcal{B})) \cdot top_1(\mathcal{B}) + q
    \end{equation}
    
    \item \textbf{Different slope and intercept} 
    \begin{equation}
         accuracy(\mathcal{B}) = (m + \Delta m \cdot pretrain(\mathcal{B})) \cdot top_1(\mathcal{B}) + (q + \Delta q \cdot pretrain(\mathcal{B}))
    \end{equation}
\end{itemize}

To compare the different models we use the goodness-of-fit ($R^2$) that represents the fraction of explained variance. Given the observed values $(x_1, y_1), \ldots, (x_N, y_N)$ (where $x_i$ is a vector of independent variables, while $y_i$ is the associated dependent variable) and the values predicted by the model $f$: $\hat{y}_1 = f(x_1), \ldots, \hat{y}_N = f(x_N)$ we can compute the mean of the observed data as:
\begin{equation}
    \bar{y} = \frac{1}{N}\sum_{i=1}^N y_i
\end{equation}
the \textit{total sum of squares} is defined as:
\begin{equation}
    SS_{tot} = \sum_{i=1}^N (y_i - \bar{y})^2
\end{equation}
while the \textit{residual sum of squares}:
\begin{equation}
    SS_{res} = \sum_{i=1}^N (y_i - \hat{y}_i)^2
\end{equation}
then the coefficient of determination can be defined as:
\begin{equation}
    R^2 = 1 - \frac{SS_{res}}{SS_{tot}}
\end{equation}

Then, even if in our case the difference is marginal, we use the adjusted $R^2$ (indicated as $\bar{R}^2$) to account for the different number of independent variables in the models, that is defined as:

\begin{equation}
    \bar{R}^2 = 1 - (1 - R^2) * \frac{N - 1}{N - df}
\end{equation}

where $df$ are the degrees-of-freedom of the model.

\begin{figure}[h!]
\begin{center}

\includegraphics[width=0.6\textwidth]{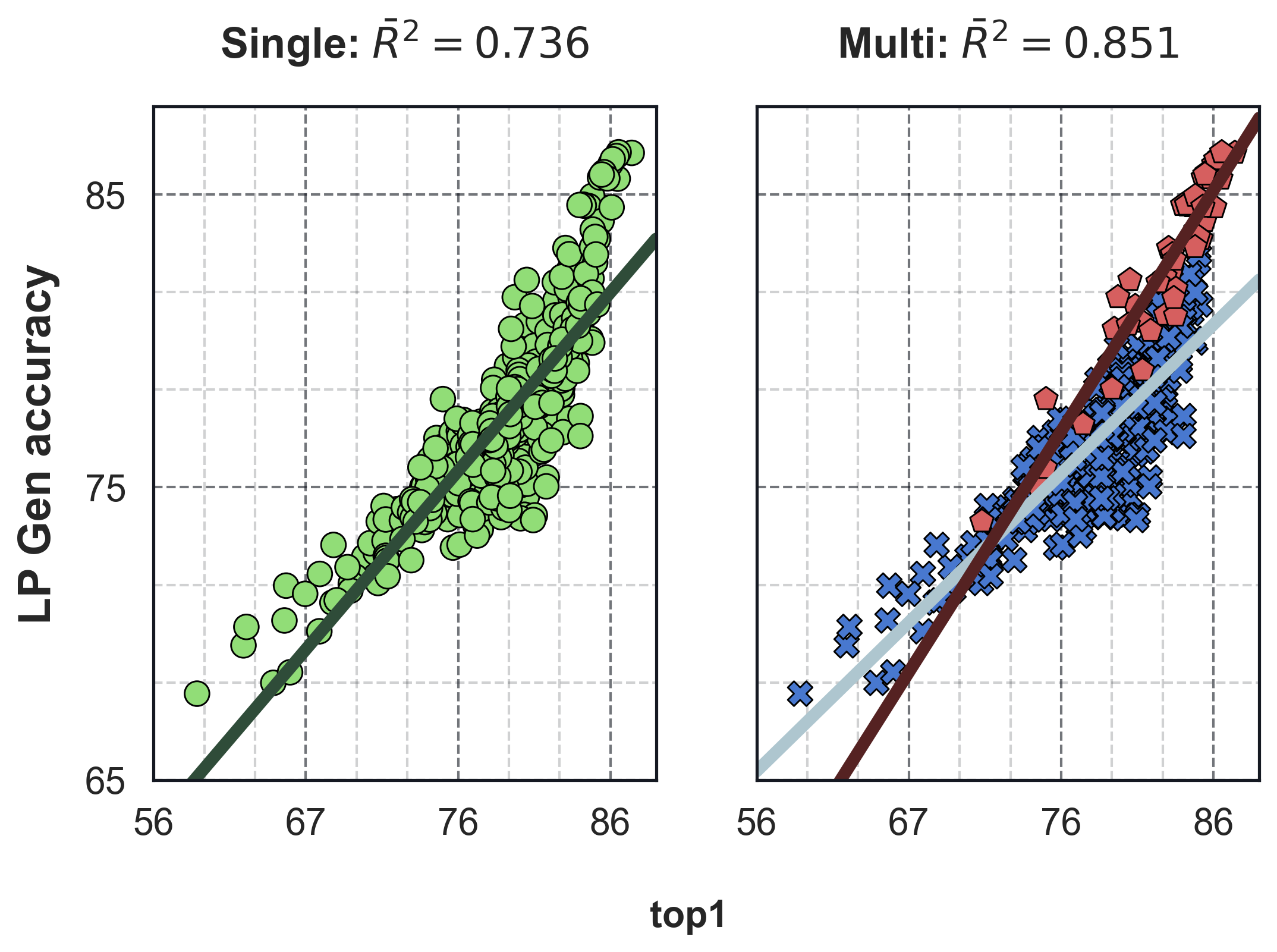}
\end{center}
\caption{Average Linear Probing generalization accuracy (over 23 domains). \label{fig:LP-gen}}
\end{figure}

\begin{figure}[h!]

\begin{center}
\includegraphics[width=0.6\textwidth]{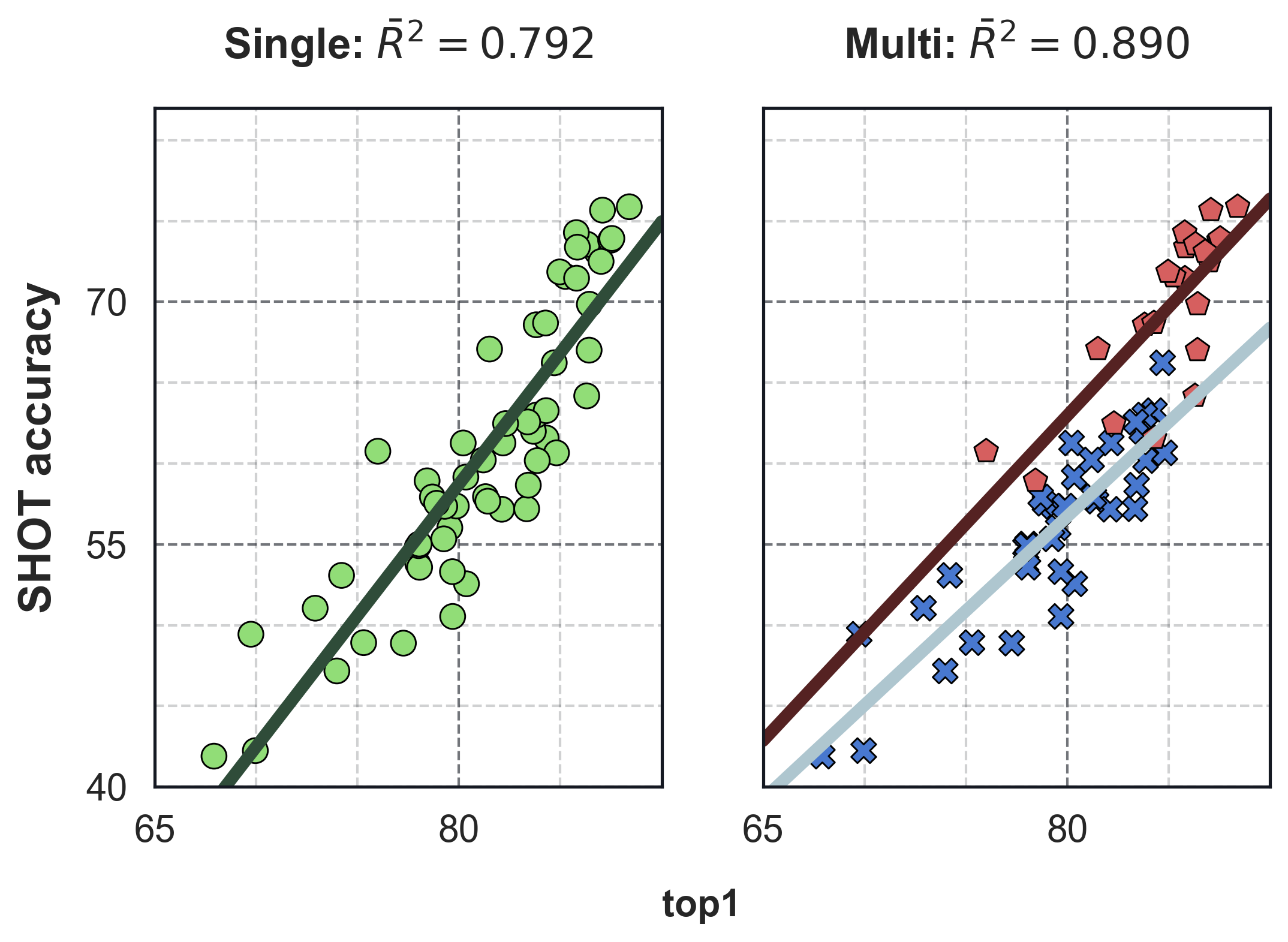}
\end{center}
\caption{Average accuracy (over 74 domain shifts) of models after applying SHOT (without fine-tuning).}
\label{shotlines}
\end{figure}

\begin{figure}[h!]
\begin{center}

\includegraphics[width=0.6\textwidth]{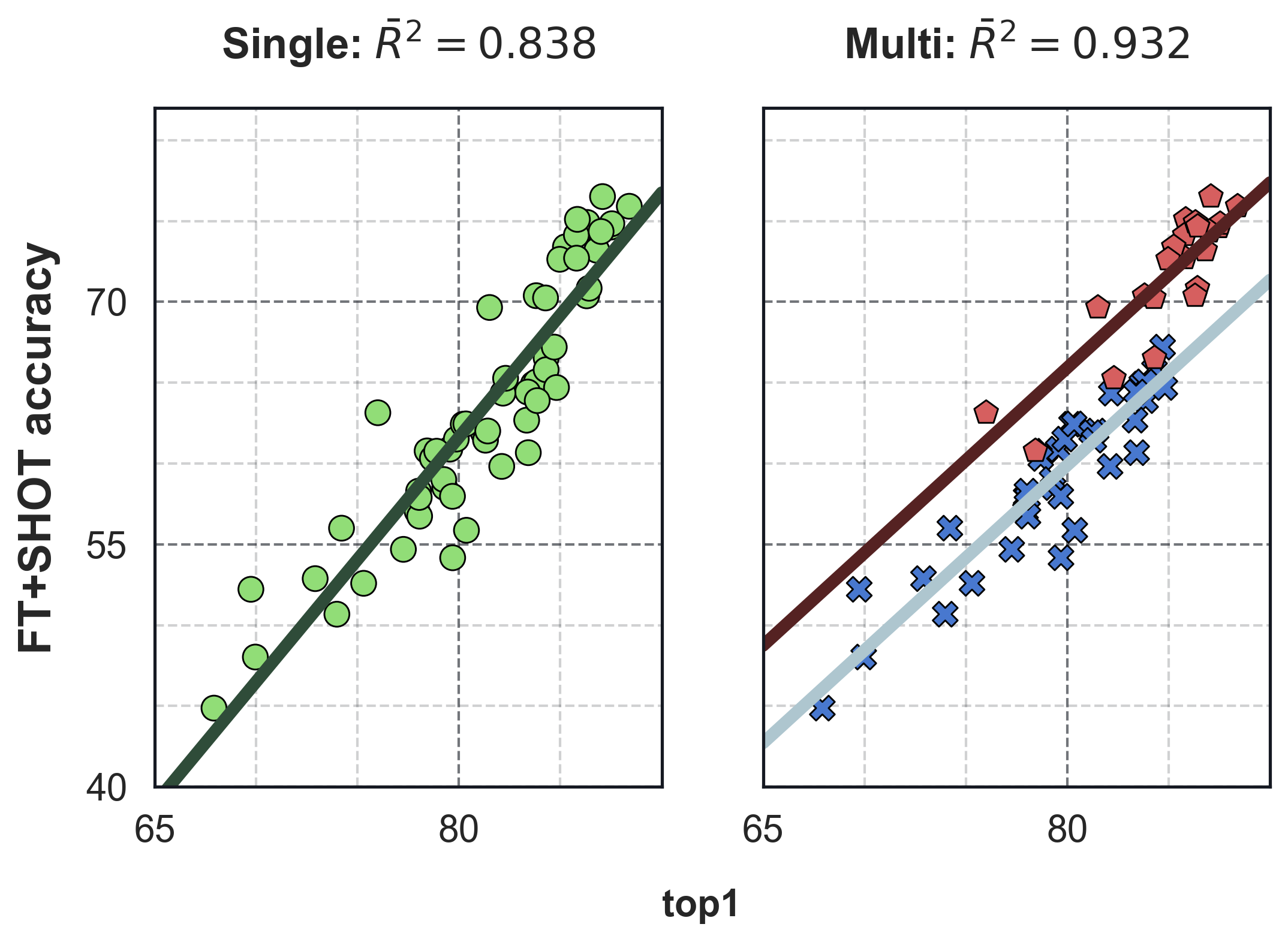}
\end{center}
\caption{Average accuracy (over 74 domain shifts) of models after applying the FT and SHOT. In this case the two lines of the statistical multi-linear model are parallel ($\Delta m$ was not statistically \label{fig:ft-shot-dgen}}
\end{figure}

We report all the coefficients of the linear and multi-linear models and the $\bar{R}^2$ values in Tab.~\ref{coefficients_table}. An illustration is given in Figs.~\ref{fig:LP-gen}-\ref{fig:ft-shot-dgen}. In some cases (for example in SHOT), we get the p-values for the $\Delta m$ and $\Delta q$ parameters higher than the statistical significance of 0.01. In this case, we just need to remove one of the two parameters to make remaining ones all significant. This phenomenon is due to the fact that the two lines, in these experiments, are almost parallel and so, $\Delta m$ is approximately 0 (see Fig.~\ref{shotlines}) and can be safely be removed from the statistical model without decreasing the goodness-of-fit.

\begin{table}[h!]
\caption{Coefficients and $\bar{R}^2$ obtained with single and multi linear models. Parameters of the multi-linear model with p-value higher than 0.01 are removed from the model, since their contribution is not statistically significant.}
\label{coefficients_table}
\begin{center}
\begin{tabular}{c|l|cccc|c}
\toprule
\bf{Experiment} & \bf{model} & $\mathbf{m}$ & $\mathbf{\Delta m} $ & $\mathbf{\Delta q}$ & $\mathbf{q}$ & $\mathbf{\bar{R}^2}$\\ 

\midrule
\multirow{2}*{LP Gen} & single reg. & $0.61$ & --- & --- & $0.29$ & $0.74$\\
& multiple reg.  & $0.51$ & $0.31$ & $-0.23$ & $0.36$ & $0.85$\\ 

\midrule
\multirow{2}*{CP Gen} & single reg. & $0.72$ & --- & --- & $0.16$ & $0.76$\\
& multiple reg. & $0.59$ & $0.23$ & $-0.14$ & $0.26$ & $0.91$\\ 

\midrule
\multirow{2}*{LP DGen} & single reg. & $1.14$ & --- & --- & $-0.41$ & $0.81$\\
& multiple reg. & $0.95$ & $0.62$ & $-0.45$ & $-0.26$ & $0.93$\\ 

\midrule
\multirow{2}*{CP DGen} & single reg. & $1.14$ & --- & --- & $-0.41$ & $0.75$\\
& multiple reg. & $0.92$ & $0.54$ & $-0.38$ & $-0.25$ & $0.92$\\ 

\midrule
\multirow{2}*{SCA} & single reg.  & $1.08$ & --- & --- & $-0.32$ & $0.73$\\
& multiple reg. & $0.87$ & $0.42$ & $-0.27$ & $-0.16$ & $0.90$\\ 
\midrule

\multirow{2}*{SHOT} & single reg. & $1.62$ & --- & --- & $-0.71$ & $0.79$\\
& multiple reg. & $1.21$ & --- & $0.07$ & $-0.40$ & $0.89$\\ 
\midrule

\multirow{2}*{FT Gen} & single reg. & $0.62$ & --- & --- & $0.35$ & $0.80$\\
& multiple reg. & $0.48$ & --- & $0.02$ & $0.46$ & $0.88$\\ 
\midrule

\multirow{2}*{FT DGen} & single reg. & $1.68$ & --- & --- & $-0.82$ & $0.70$\\
& multiple reg. & $1.16$ & --- & $0.09$ & $-0.43$ & $0.82$\\ 
\midrule

\multirow{2}*{FT+SCA} & single reg. & $1.62$ & --- & --- & $-0.72$ & $0.67$\\
& multiple reg. & $1.10$ & --- & $0.08$ & $-0.33$ & $0.80$\\ 
\midrule

\multirow{2}*{FT+SHOT} & single reg. & $1.51$ & --- & --- & $-0.59$ & $0.84$\\
& multiple reg. & $1.15$ & --- & $0.06$ & $-0.32$ & $0.93$\\ 

\bottomrule
\end{tabular}
\end{center}
\end{table}

\begin{figure}[h!]
\begin{center}
\includegraphics[width=0.6\textwidth]{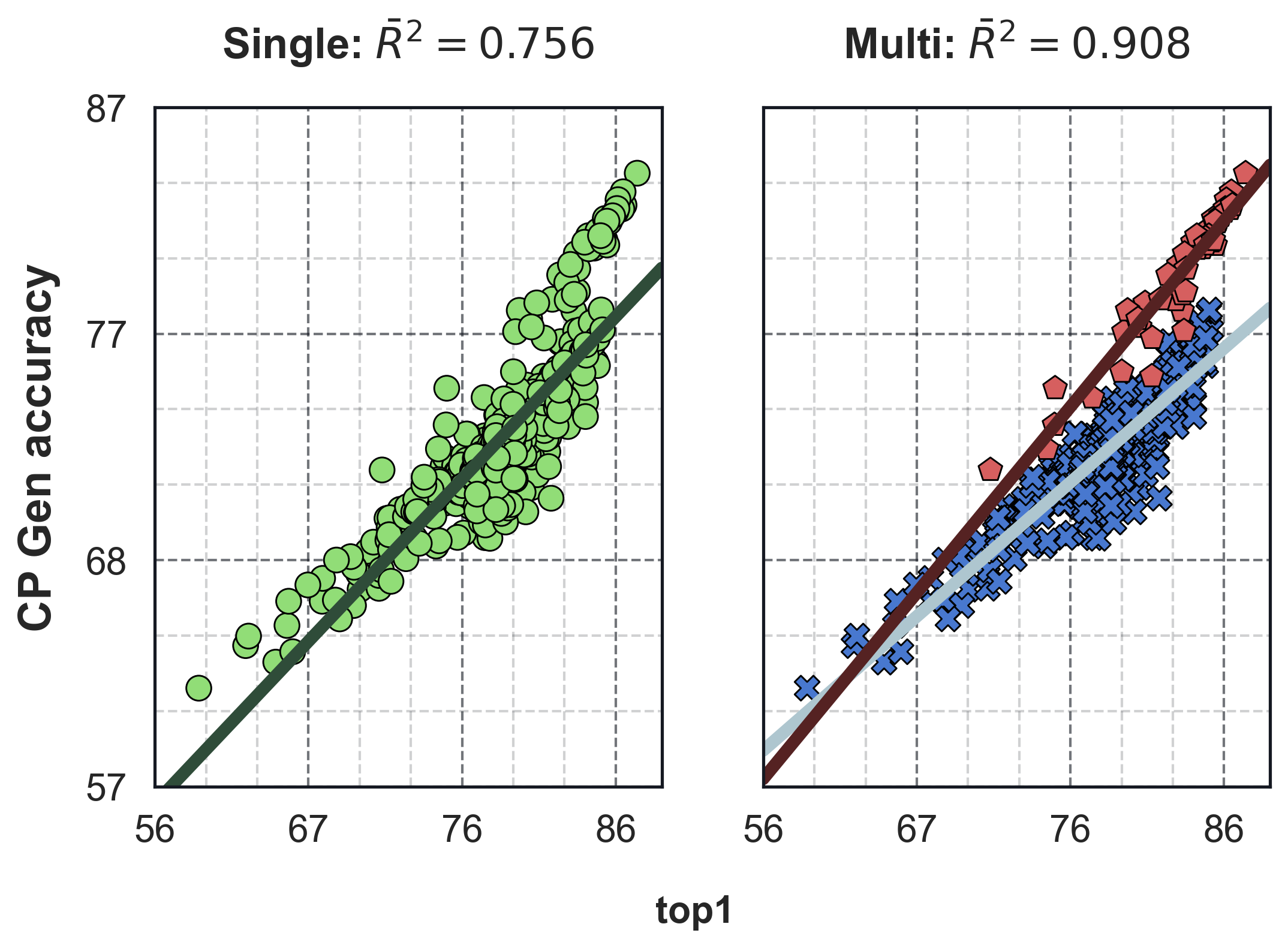}
\end{center}
\caption{Average Cluster Probing generalization accuracy (over 23 domains).}
\end{figure}

\begin{figure}[h!]
\begin{center}
\includegraphics[width=0.6\textwidth]{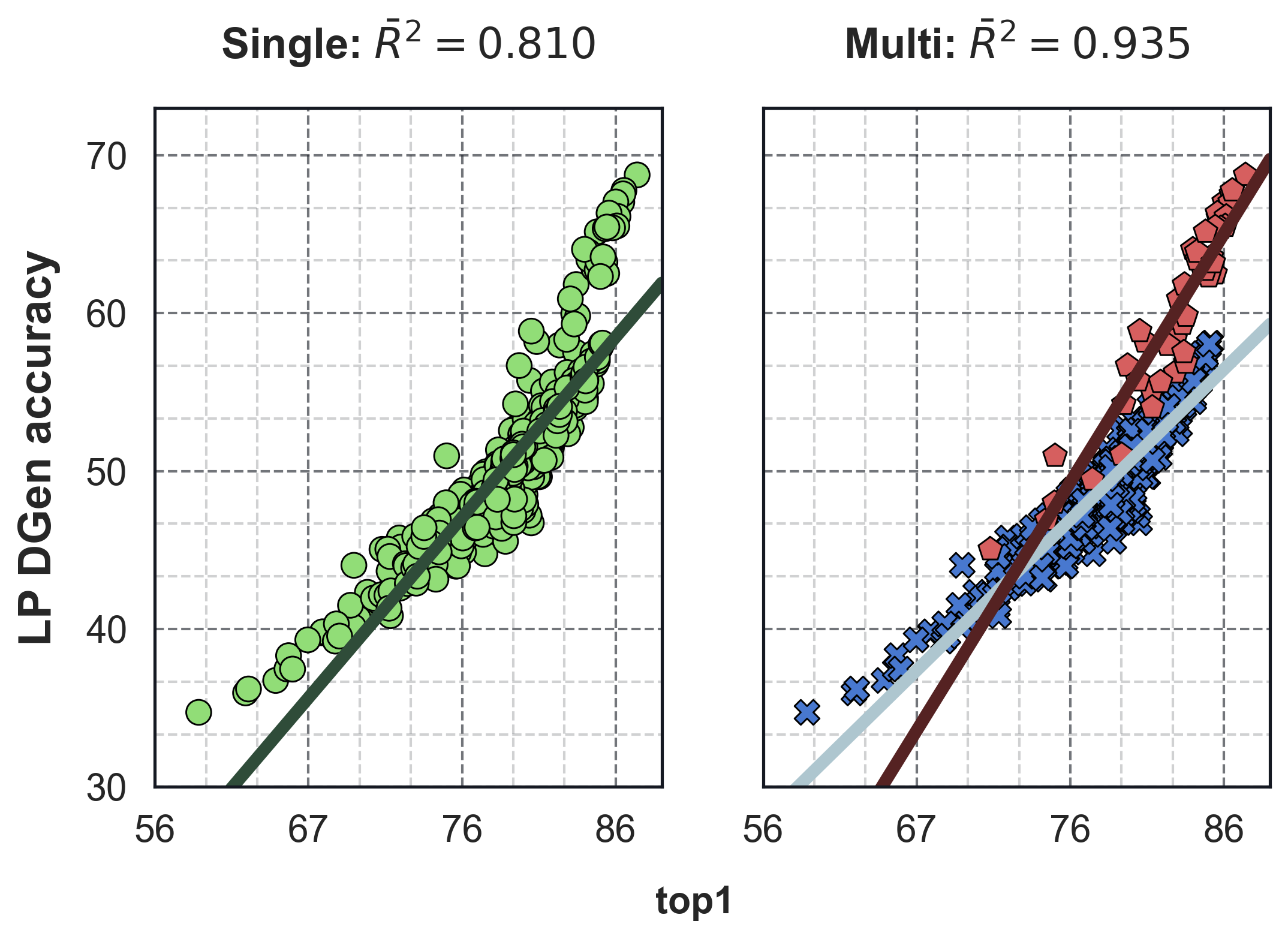}
\end{center}
\caption{Average Linear Probing domain generalization accuracy (over 74 domain shifts).}
\end{figure}

\begin{figure}[h!]
\begin{center}
\includegraphics[width=0.6\textwidth]{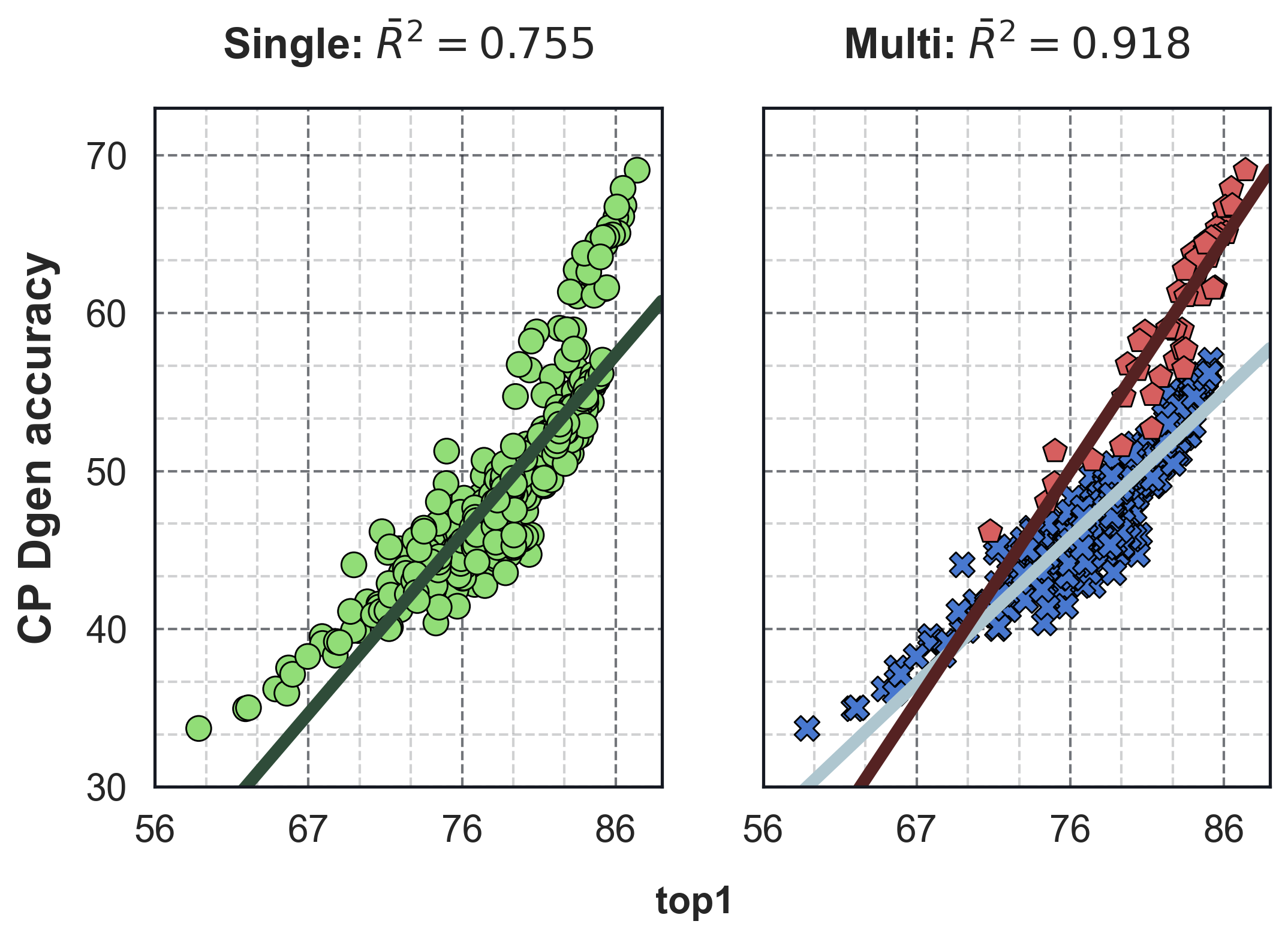}
\end{center}
\caption{Average Cluster Probing domain generalization accuracy (over 74 domain shifts).}
\end{figure}

\begin{figure}[h!]
\begin{center}
\includegraphics[width=0.6\textwidth]{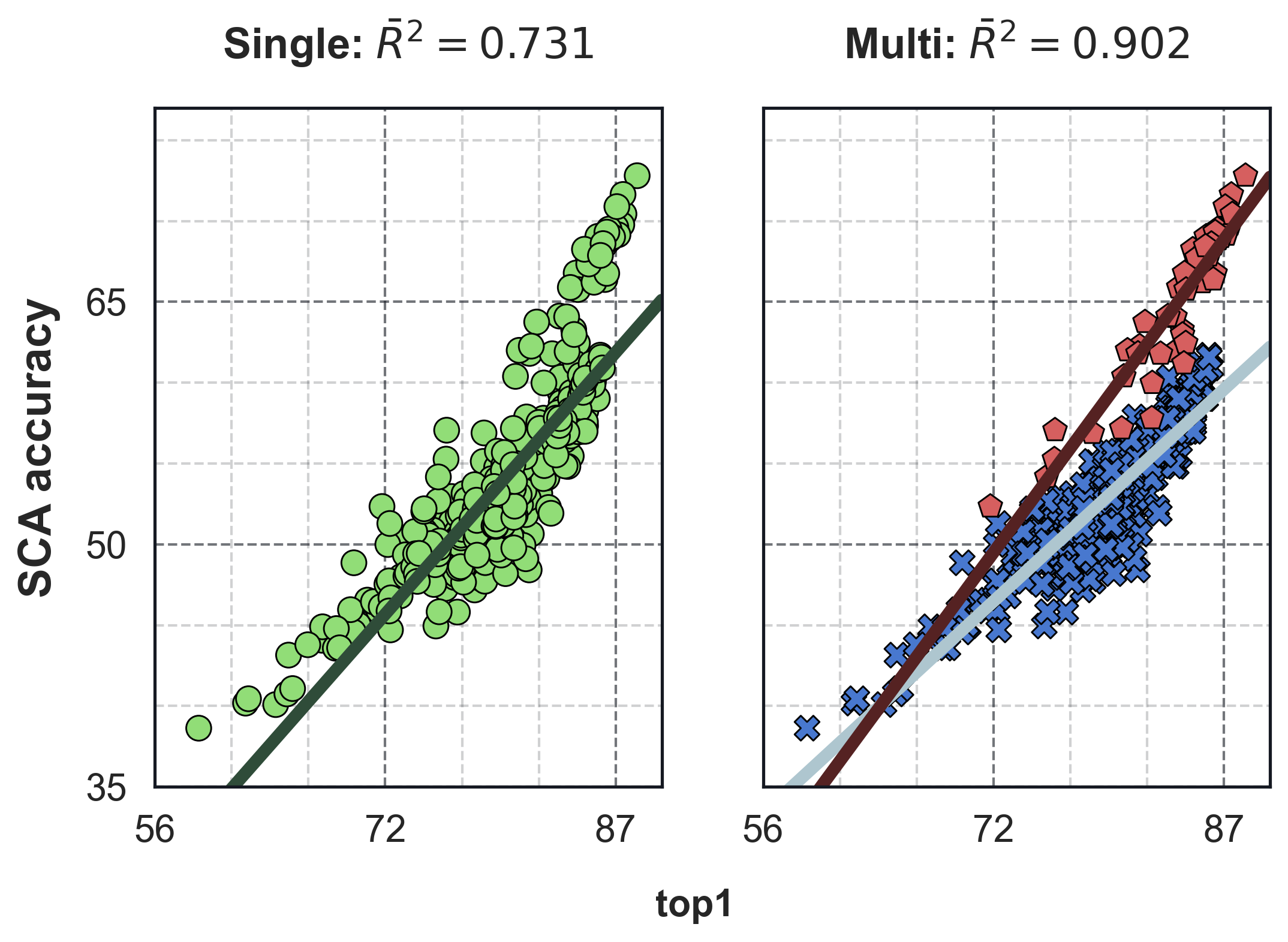}
\end{center}
\caption{Average accuracy (over 74 domain shifts) of models after applying SCA (without fine-tuning).}
\end{figure}

\begin{figure}[h!]
\begin{center}
\includegraphics[width=0.6\textwidth]{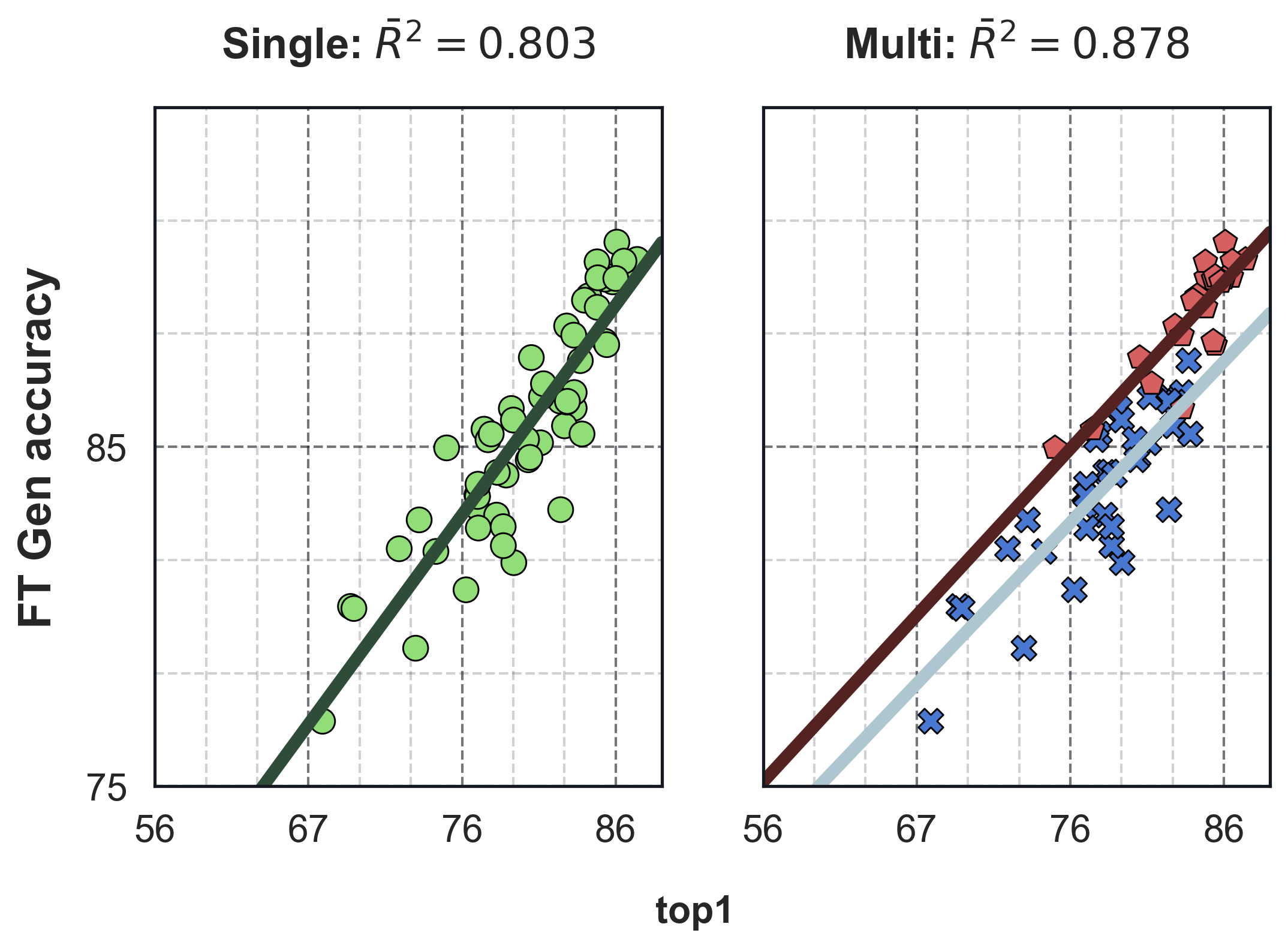}
\end{center}
\caption{Average generalization accuracy (over 23 domains) of models after applying the FT.}
\end{figure}

\begin{figure}[h!]
\begin{center}
\includegraphics[width=0.6\textwidth]{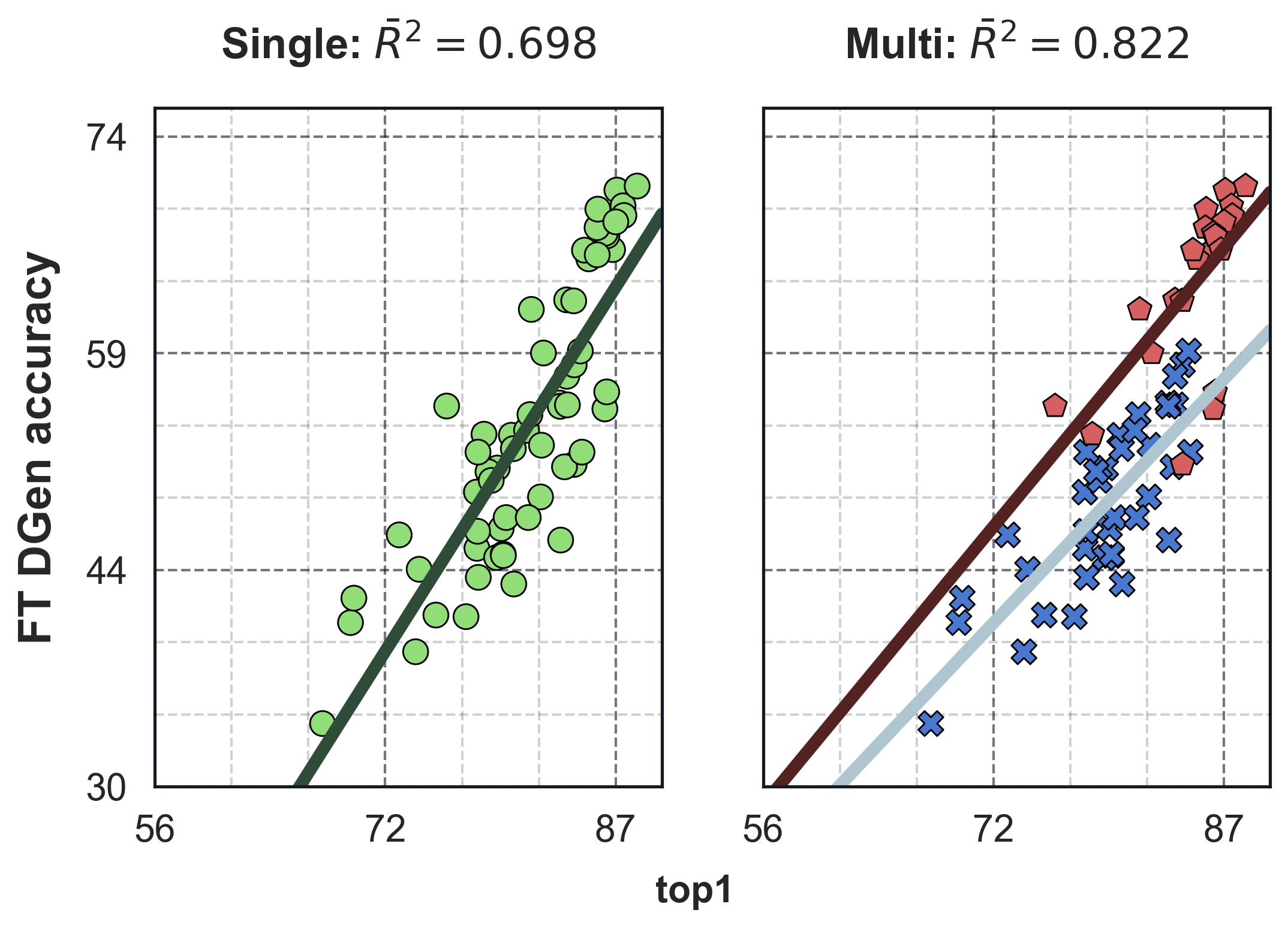}
\end{center}
\caption{Average domain generalization accuracy (over 74 domain shifts) of models after applying the FT.}
\end{figure}

\begin{figure}[h!]
\begin{center}
\includegraphics[width=0.6\textwidth]{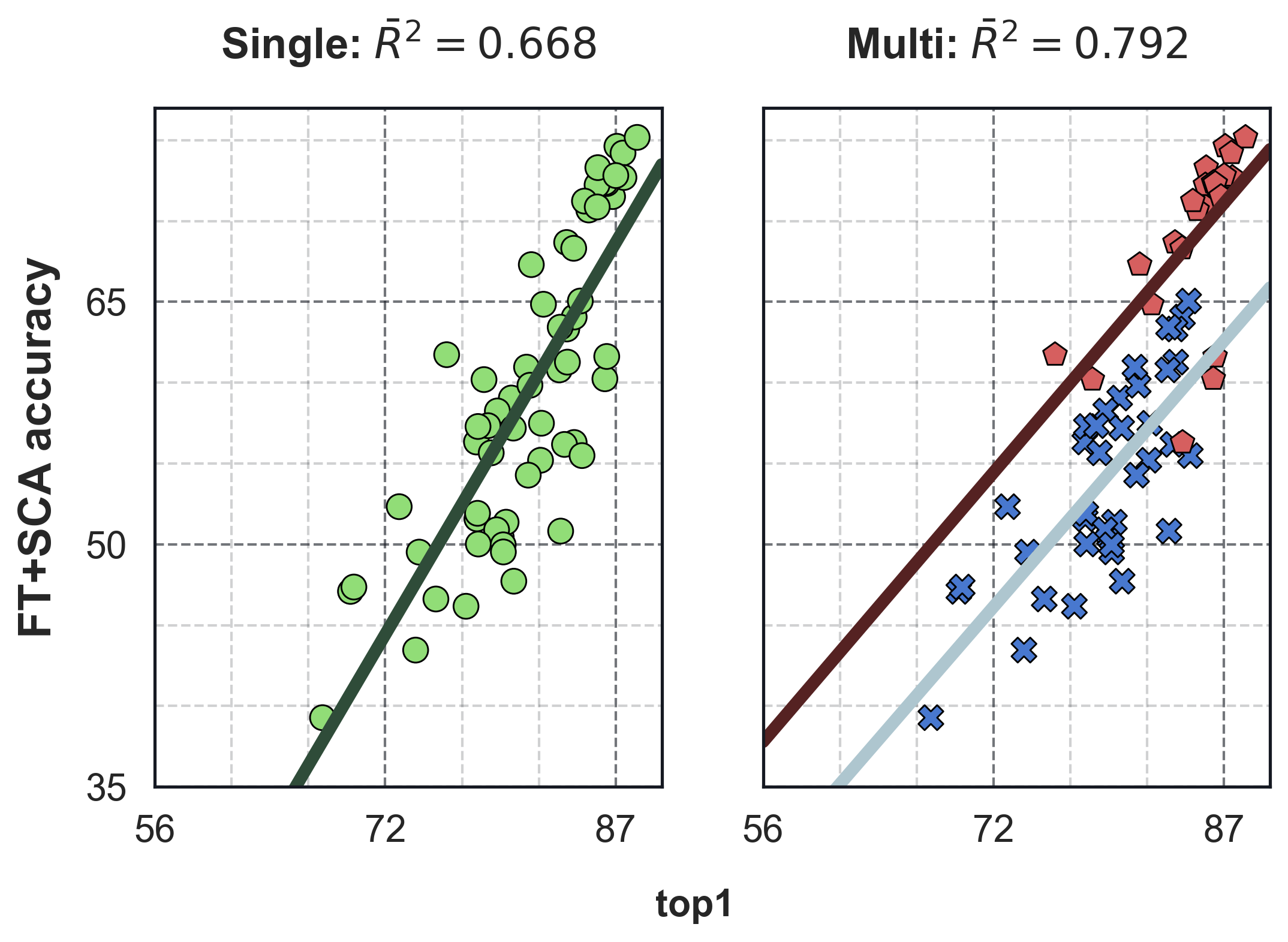}
\end{center}
\caption{Average accuracy (over 74 domain shifts) of models after applying the FT and SCA.}
\end{figure}

\clearpage

\section{Comparison of Domain Generalization and SCA with semantic pre-training}
\label{semantictrain}

We further investigate the role of pre-training strategy. Tab.~\ref{tab:miil-naive} contrasts the domain generalization accuracy for miil ~\citep{ridnik2021imagenet} and naive pre-training on different domain generalization datasets. Tab.~\ref{tab:timm-miil-naive} shows results for different initializations of ResNet50 on Office31. 
\begin{table}[h!]
\begin{center}
    \begin{threeparttable}
\caption{Comparison of results (accuracies \%) of semantic pre-training (miil) of ~\cite{ridnik2021imagenet} and naive pre-training of ResNet50, for linear probing domain generalization and SCA. The results are avaraged across the domains of the datasets.} 
\vspace{0.5cm}
\label{tab:miil-naive}

\begin{footnotesize}
\begin{tabular}{llcccccc}
\toprule
\textbf{Experiment} & \textbf{Pre-train} & \textbf{O31} &  \textbf{Visda} &  \textbf{O.Home} &  \textbf{Adapt.} &  \textbf{I-CLEF} &  \textbf{D.Net} \\
\midrule
\multirow{2}{*}{LP DGen} & naive\tnote{\textdagger}     &  86.6 &   63.1 &        68.5 &       59.1 &     79.8 &       19.9 \\
                      & miil                         &  88.8 &   67.3 &        71.4 &       68.2 &     81.9 &       25.3 \\
\midrule
\multirow{2}{*}{SCA}  & naive\tnote{\textdagger}     &  90.3 &   80.7 &        74.4 &       73.8 &     85.1 &       23.6 \\
                      & miil                         &  90.4 &   79.8 &        72.2 &       73.3 &     84.8 &       21.3 \\

\bottomrule
\end{tabular}
\end{footnotesize}

    \begin{tablenotes}
        \footnotesize
        \item[\textdagger] Performed by us.
    \end{tablenotes}
    
    \end{threeparttable}
        \end{center}
\end{table}

\begin{table}[h!]
\caption{Comparison of results (accuracies \%) on Office31 benchmark using different pre-trained weights of Resnet50. We included the following pre-training: \textbf{timm}~\citep{rw2019timm}, the semantic training of~\cite{ridnik2021imagenet} (\textbf{miil}) and our naive pretrain. Office31 domains are \textbf{A}mazon, \textbf{W}ebcam and \textbf{D}SLR and the experiments are indicated with \textbf{source} $\to$ \textbf{target}. For these experiments the official code (not distributed and without Automatic Mixed Precision) of SHOT~\citep{liang2020we} and NRC~\citep{yang2021exploiting} have been used, so the results may be slightly different from the ones reproduced with our implementation.} 
\label{tab:timm-miil-naive}
\begin{center}
\begin{adjustbox}{max width=0.9\textwidth}
\begin{tabular}{ccccccccc|c}

 \multicolumn{10}{c}{\textbf{Office31 (ResNet50)}} \\
\toprule
\textbf{Method} & \textbf{Init} & \textbf{IN21k} & \textbf{A $\to$ W} & \textbf{A $\to$ D} & \textbf{W $\to$ A} & \textbf{W $\to$ D} & \textbf{D $\to$ A} & \textbf{D $\to$ W} & \textbf{Avg}\\
\midrule
\multirow{3}{*}{LP DGen} & timm & & 77.5 & 78.9 & 68.3 & 97.8 & 64.6 & 96.6 & 80.6 \\
                         & miil & \checkmark & 89.4 & 89.1 & 78.3 & 99.4 & 78.7 & 98.1 & 88.8 \\
                         & naive & \checkmark & 86.2 & 83.3 & 77.3 & 99.4 & 76.4 & 97.1 & 86.6 \\

\midrule
\multirow{3}{*}{SCA} & timm & & 89.2 & 92.5 & 73.4 & 97.2 & 73.9 & 90.1 & 86.0 \\
                     & miil & \checkmark & 93.8 & 97.4 & 79.0 & 98.2 & 78.8 & 95.2 & 90.4 \\
                     & naive & \checkmark & 93.3 & 94.6 & 80.1 & 98.2 & 80.0 & 95.7 & 90.3 \\

\midrule
\multirow{3}{*}{FT+SHOT} & timm & & 89.3 & 87.1 & 68.9 & 99.8 & 67.1 & 98.1 & 85.5 \\
                      & miil & \checkmark & 93.2 & 92.4 & 75.3 & 99.8 & 76.8 & 98.8 & 89.4 \\
                      & naive & \checkmark & 91.1 & 94.0 & 75.3 & 99.6 & 73.5 & 98.8 & 88.7 \\

\midrule
\multirow{3}{*}{FT+NRC} & timm & & 94.1 & 93.0 & 72.2 & 99.8 & 69.8 & 98.2 & 87.7 \\
                     & miil & \checkmark & 96.0 & 89.6 & 76.7 & 99.6 & 75.0 & 98.1 & 89.2 \\
                     & naive & \checkmark & 92.6 & 96.0 & 80.4 & 99.8 & 78.5 & 98.4 & 90.9 \\
\bottomrule
\end{tabular}
\end{adjustbox}
\end{center}
\end{table}

\clearpage

\section{Batch Normalization vs Layer Normalization}
\label{app_ln_vs_bn}

\begin{figure}[h!]
\begin{center}
\includegraphics[width=0.99\textwidth]{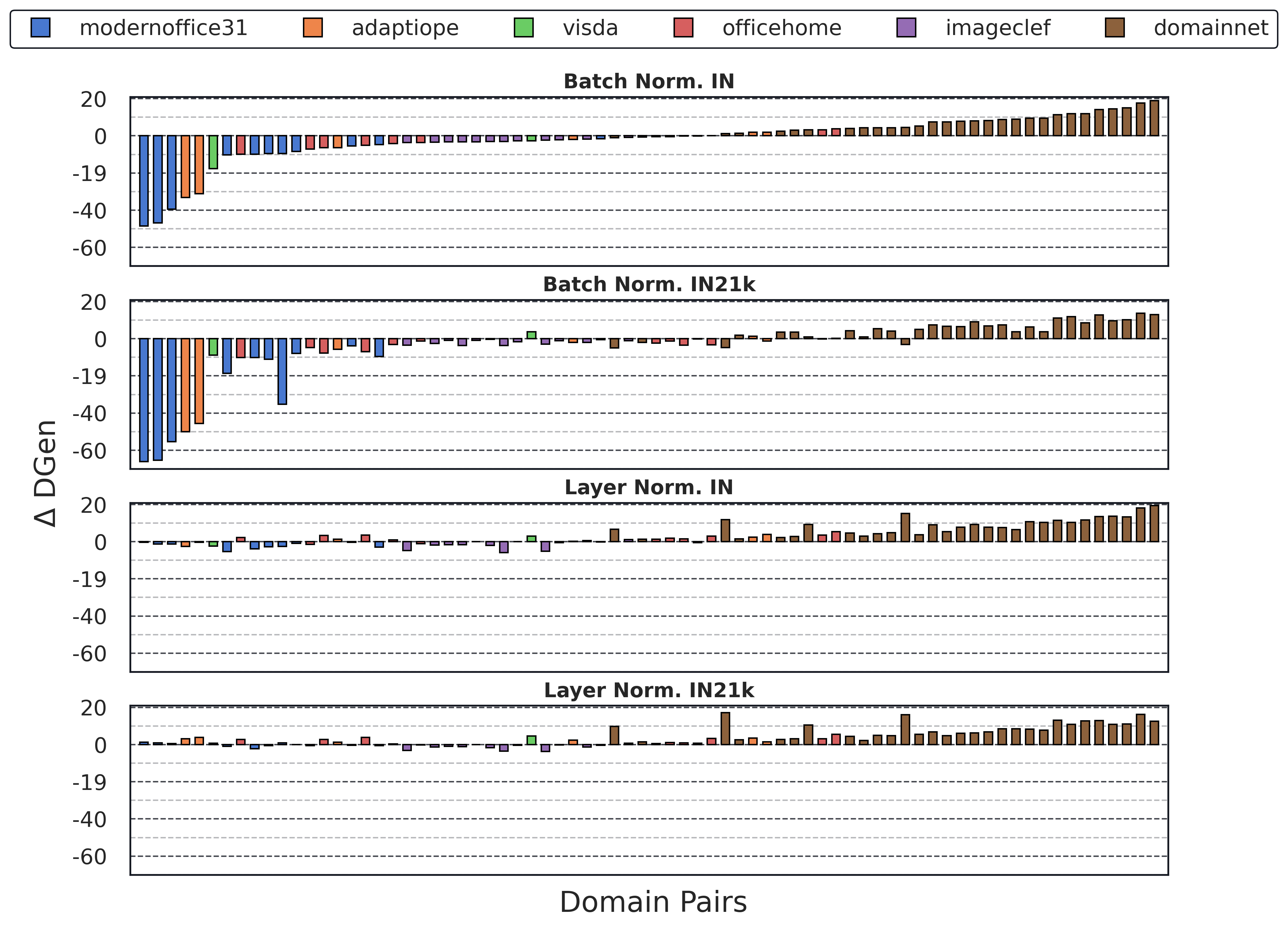}
\end{center}
\caption{Each bar represent a different domain-shift experiment, i.e., a source and a target. Bars height are the average gain/degradation of domain generalization (always w.r.t. the baseline LP domain generalization) for the given domain pair.}
\label{domains_bn_ln}
\end{figure}

Fixing a source-target domain pair, we compute the average domain generalization improvement/degradation over the different architectures. In particular, we split all 59 architectures considered into 4 groups based on normalization layers and pre-training: BN+IN (24 models), BN+IN21k (8 models), LN+IN (8 models) and LN+IN21k (19 models).
In Fig.~\ref{domains_bn_ln} we show the average improvement/degradation for each domain pair and for every model group. As it is possible to see, there are some domain pairs where the models with BN degrades a lot the domain generalization, while all models with LN are more stable.

In Tab.~\ref{degradation_table} are reported the numerical results for the domain pairs with the highest degradation caused by the fine-tuning with Batch Normalization layers. In particular, the domain pairs correspond to the 10 left-most bars of Fig.~\ref{domains_bn_ln}

\begin{table}[h]
\caption{Domain pairs with the highest fine-tuning domain generalization degradation caused by Bacth Normalization layers.}
\vspace{0.5cm}
\label{degradation_table}
\begin{center}
\begin{footnotesize}

\begin{tabular}{lllrrrr}
\toprule
       dataset &    source &         target &  bn\_in1k &  bn\_in21k &  ln\_in1k &  ln\_in21k \\
\midrule
modernoffice31 & synthetic &         webcam &    -48.6 &     -66.1 &     -0.3 &       1.2 \\
modernoffice31 & synthetic &           dslr &    -46.8 &     -65.4 &     -1.3 &       1.0 \\
modernoffice31 & synthetic &         amazon &    -39.5 &     -55.4 &     -1.3 &       0.6 \\
     adaptiope & synthetic &      real\_life &    -33.1 &     -50.0 &     -2.6 &       3.2 \\
     adaptiope & synthetic & product\_images &    -31.2 &     -45.7 &     -0.3 &       3.9 \\
         visda &     train &     validation &    -17.6 &      -8.7 &     -2.4 &       0.8 \\
modernoffice31 &      dslr &      synthetic &    -10.3 &     -18.5 &     -5.3 &      -0.9 \\
    officehome &   Clipart &            Art &    -10.0 &     -10.2 &      2.3 &       2.7 \\
modernoffice31 &    webcam &      synthetic &    -10.0 &     -10.0 &     -3.9 &      -2.2 \\
modernoffice31 &    webcam &         amazon &     -9.6 &     -11.0 &     -2.7 &      -0.5 \\
\bottomrule
\end{tabular}
\end{footnotesize}
\end{center}

\end{table}

In Tab.~\ref{degradation_avg} we report the average degradation in case of failure on all domain pairs for different model groups. Also here it is possible to see that the models with Batch Normalization are affected from high degradation of the domain generalization with the fine-tuning.

\begin{table}[h]
\caption{Average degradation (\%) in case of failure (averaged across 74 domain pairs). Every row select a certain type of model to make a full comparison. Between parethesis in reported the number of model in that group. } 
\vspace{0.5cm}
\label{degradation_avg}
\begin{center}
\begin{footnotesize}
\begin{tabular}{l ccccc}

\toprule
\textbf{Models} & \textbf{FT DGen} &  \textbf{SCA} & \textbf{SHOT} & \textbf{FT+SCA} & \textbf{FT+SHOT} \\
\midrule
\textbf{All} (59) 
& -6.67  \tiny{$\pm$  4.55} 
& -1.55  \tiny{$\pm$ 0.67} 
& -3.11  \tiny{$\pm$ 2.76} 
&  -7.88 \tiny{$\pm$ 6.75} 
& -1.58  \tiny{$\pm$ 2.04}\\
\midrule
\textbf{IN} (32) 
& -8.05  \tiny{$\pm$ 3.57} 
& -1.79  \tiny{$\pm$ 0.68} 
& -2.66  \tiny{$\pm$ 1.38} 
&  -8.79 \tiny{$\pm$ 5.75} 
& -1.36  \tiny{$\pm$ 0.91} \\
\textbf{IN21K} (27) 
& -5.02  \tiny{$\pm$ 5.07} 
& -1.27  \tiny{$\pm$ 0.54} 
& -3.64  \tiny{$\pm$ 3.76} 
&  -6.65 \tiny{$\pm$ 7.86} 
& -1.89  \tiny{$\pm$ 2.97}\\
\midrule
\textbf{BN} (32) 
& -10.51   \tiny{$\pm$ 2.22} 
& -1.92    \tiny{$\pm$ 0.63} 
& -2.83    \tiny{$\pm$ 1.34} 
&  -12.72  \tiny{$\pm$ 4.89} 
& -1.82    \tiny{$\pm$ 2.46} \\
\textbf{LN} (27) 
& -2.10   \tiny{$\pm$ 0.65} 
& -1.12   \tiny{$\pm$ 0.40} 
& -3.43   \tiny{$\pm$ 3.82} 
&  -1.41  \tiny{$\pm$ 0.77} 
& -1.27   \tiny{$\pm$ 1.26} \\
\midrule
\textbf{IN+BN} (24) 
& -9.82    \tiny{$\pm$ 2.01} 
& -1.95    \tiny{$\pm$ 0.70} 
& -2.56    \tiny{$\pm$ 0.98} 
&  -11.26  \tiny{$\pm$ 4.37} 
& -1.27    \tiny{$\pm$ 0.72}\\
\textbf{IN+LN} (8) 
& -2.75  \tiny{$\pm$ 0.63} 
& -1.32  \tiny{$\pm$ 0.34} 
& -2.96  \tiny{$\pm$ 2.26} 
&  -1.39 \tiny{$\pm$ 0.57} 
& -1.60  \tiny{$\pm$ 1.33} \\
\midrule
\textbf{IN21k+BN} (8) 
& -12.60  \tiny{$\pm$ 1.40} 
& -1.82   \tiny{$\pm$ 0.40} 
& -3.65   \tiny{$\pm$ 1.94} 
&  -17.10 \tiny{$\pm$ 3.75} 
& -3.31   \tiny{$\pm$ 4.47}\\
\textbf{IN21k+LN} (19) 
& -1.83 \tiny{$\pm$ 0.43} 
& -1.03 \tiny{$\pm$ 0.40} 
& -3.63 \tiny{$\pm$ 4.35} 
&  -1.42\tiny{$\pm$ 0.87} 
& -1.08 \tiny{$\pm$ 1.22} \\
\bottomrule

\end{tabular}
\end{footnotesize}
\end{center}
\end{table}

\clearpage

\section{Computational times.}
\label{comptimes}

In Tab.~\ref{times} we report the average number of seconds to perform SCA, one epoch of fine-tuning and one epoch of SHOT adaptation for Office31 and Visda datasets (averaged across all different domains).
We remark that SCA is performed just once, while the fine-tuning (in our experiments) uses between 15 (10 + 5, see technical details in App.~\ref{techdet}) and 100 epochs, while SHOT uses always 15 epochs.

\begin{table}[h!]
    \centering
\caption{Time required (in seconds) for fine-tuning and adaptation. The times are averaged on the domains of the used dataset. For fine-tuning and SHOT the time reported refers to one epoch.We remark that according to standard training, SHOT requires 15 epochs to adapt while DANN~\citep{ganin2016domain} requires 100 epochs. Models are in mixed precision training with batch size 64 split into 2 $\times$ NVIDIA RTX 2070 SUPER on a single machine with AMD Ryzen 3900x.} 
\vspace{0.5cm}
\label{times}
\begin{tabular}{cccccc}
\toprule
\textbf{Dataset} & \textbf{Model} &\textbf{FT} & \textbf{SCA} & \textbf{SHOT}  & DANN \\
\midrule
\multirow{2}*{Office31} & ResNet50 & 3.4 & 1.7 & 6.7 & 7.0\\
\cmidrule(lr){2-6}
& ViT Base & 5.7 & 4.5 & 14.8 & 10.8\\
\midrule
\multirow{2}*{Visda} & ResNet50 & 211.1 & 92.1 & 425.3 & 404.1\\
\cmidrule(lr){2-6}
& ViT Base & 381.6 & 327.1 & 1068.5 & 790.3 \\
\bottomrule
\end{tabular}
\end{table}

\section{Additional results: best performing models.}
\label{sec:additional_results}
In Tab.~\ref{additional_results} we provide domain generalization results on large architectures on different DA datasets when fine-tuning varying on the fine-tuning choice.

\begin{table}[h]
\centering
\caption{Domain generalization accuracy for large backbones with different fine-tuning decisions. Both the fine-tuning and non fine-tuning cases are considered. For each architecture, the accuracy change is the average over the domains of the considered dataset.}

\vspace{0.5cm}

\label{additional_results}

\begin{adjustbox}{max width=0.9\textwidth}
\begin{tabular}{l|lc|ccccccc}

\toprule
\textbf{Model} &   \textbf{Method} &  \textbf{FT} & \textbf{Office31} &  \textbf{Visda} &  \textbf{OfficeHome} &  \textbf{Adapt.} &  \textbf{Mod.Office31} &  \textbf{Image-CLEF} &  \textbf{DomainNet} \\
\midrule
\multirow{4}{*}{BEIT L 384}        &  \multirow{2}{*}{SCA} & \xmark     &      94.8 &   85.1 &        89.0 &       93.1 &            97.4 &       88.6 &       45.3 \\
                                   &                       & \checkmark &      94.2 &   79.6 &        88.9 &       92.7 &            95.5 &       86.4 &       45.0 \\
\cmidrule{2-10}
                                   & \multirow{2}{*}{SHOT} & \xmark     &      95.5 &   89.2 &        92.2 &       95.8 &            96.5 &       88.3 &       45.8 \\
                                   &                       & \checkmark &      94.6 &   90.7 &        90.1 &       95.1 &            96.6 &       89.6 &       45.1 \\
\midrule
\multirow{4}{*}{SWIN L 224}        &  \multirow{2}{*}{SCA} & \xmark     &       95.0 &   83.4 &        84.9 &       88.8 &            96.3 &       88.3 &       37.9 \\
                                   &                       & \checkmark &       94.4 &   83.0 &        86.4 &       88.2 &            93.3 &       86.0 &       49.1 \\
\cmidrule{2-10}
                                   & \multirow{2}{*}{SHOT} & \xmark     &       94.8 &   88.5 &        90.6 &       91.5 &            96.4 &       88.5 &       47.1 \\
                                   &                       & \checkmark &       95.0 &   88.9 &        89.3 &       91.9 &            96.0 &       88.0 &       51.2 \\
\midrule
\multirow{4}{*}{VIT L 384}         &  \multirow{2}{*}{SCA} & \xmark     &      95.3 &   82.6 &        88.0 &       89.3 &            96.4 &       88.2 &       42.4 \\
                                   &                       & \checkmark &      94.0 &   83.5 &        88.6 &       89.5 &            94.6 &       86.7 &       52.6 \\
\cmidrule{2-10}
                                   & \multirow{2}{*}{SHOT} & \xmark     &      94.9 &   89.0 &        91.5 &       93.9 &            95.4 &       89.2 &       51.5 \\
                                   &                       & \checkmark &      93.9 &   83.9 &        91.3 &       92.4 &            95.8 &       89.5 &       54.1 \\
\midrule
\multirow{4}{*}{ConvNext XL 384}   &  \multirow{2}{*}{SCA} & \xmark     &     95.5 &   84.9 &        86.9 &       90.0 &            96.5 &       88.8 &        41.2 \\
                                   &                       & \checkmark &      94.8 &   84.1 &        87.4 &       90.8 &            93.8 &       83.2 &       49.8 \\
\cmidrule{2-10}
                                   & \multirow{2}{*}{SHOT} & \xmark     &      94.0 &   89.2 &        90.3 &       93.9 &            94.0 &       89.0 &       48.5 \\
                                   &                       & \checkmark &      95.1 &   88.1 &        88.8 &       93.3 &            92.2 &       89.4 &       51.0 \\
\bottomrule

\end{tabular}
\end{adjustbox}
\end{table}


\section{Self-supervised pre-training}
\label{sec:ssl}

In all previous experiments we considered only supervised pre-training. However, recently, the community shows a great interest on self-supervised learning (SSL) techniques, which proved to achieve remarkable performance. Thus in Tab.~\ref{tab:ssl}, we present the results of some  experiments on ResNet50 trained with three different SSL methods, namely, DINO \citep{dino}, MOCO v1 \citep{mocov1} and MOCO v2 \citep{mocov2}.
As it can be noticed, after performing a fine-tuning on the source domain (and eventually applying SCA) MOCO v2 performs very similarly (on average) with a supervised pre-training on ImageNet, while DINO and MOCO v1 achieve a lower accuracy.
When FT+SHOT is applied MOCO v1 still underperforms other methods, while DINO and MOCO v2 achieve similar and promising results which, however, are still marginally below the performance of the supervised pre-training, which gets the best accuracy (66.1\%). 

From our findings and extensive experiments reported in the main text, we know that the fine-tuning on the source domain of a model with Batch Normalization layers (like ResNet50) can lead to large  drop in performance in some scenarios and the right way to adapt these models after the fine-tuning is to adapt the feature extractor (with SHOT, for example). 
Moreover, although the results are promising for SSL pre-training and even if MOCO-v2 performed similarly to supervised pre-training on FT and FT+SCA, the drop in accuracy for  FT+SHOT highlights that SSL pre-training still underperforms supervised pre-training (on ImageNet1k) on SF-UDA, when the right adaptation choices are made.

\begin{table}[h!]
\begin{center}

\caption{Target accuracy (\%) using different initialization weights of ResNet50 after performing the following experiments: fine-tuning on the source domain (FT), FT+SCA and FT+SHOT.}
\vspace{0.5cm}
\label{tab:ssl}

\begin{footnotesize}
\begin{tabular}{llcccccc||c}
\toprule
\textbf{Experiment} & \textbf{Pre-train} & \textbf{MO31} &  \textbf{Visda} &  \textbf{O.Home} &  \textbf{Adapt.} &  \textbf{I-CLEF} &  \textbf{D.Net} &  \textbf{Avg}\\
\midrule
\multirow{4}{*}{FT} & DINO     &  42.7 &   45.6 &        45.7 &       30.9 &     71.3 &       19.7 & 42.7\\
                           & MOCO v1  &  38.5 &   44.6 &        41.7 &       28.6 &     66.2 &       20.1 & 40.0\\
                           & MOCO v2  &  54.7 &   \textbf{54.2} &        55.1 &       39.5 &     \textbf{76.9} &       \textbf{25.5} & \textbf{51.0}\\
                           & Supervised (IN1k)  &  \textbf{56.4} &   49.6 &        \textbf{55.5} &       \textbf{43.3} &     75.9 &       21.6 & 50.4\\
\midrule
\multirow{4}{*}{FT+SCA}    & DINO     &  50.8 &   54.5 &        49.8 &       35.9 &     75.0 &       21.9 & 48.0\\
                           & MOCO v1  &  49.5 &   52.3 &        43.9 &       34.3 &     70.4 &       23.9 & 45.7\\
                           & MOCO v2  &  \textbf{63.4} &   \textbf{56.9} &        55.8 &       45.3 &     \textbf{79.5} &       \textbf{28.0} & 54.8\\
                           & Supervised (IN1k) &  62.7 &   55.5 &        \textbf{59.4} &       \textbf{49.6} &     79.3 &       23.6 & \textbf{55.0}\\
\midrule
\multirow{4}{*}{FT+SHOT}   & DINO     &  71.8 &   67.3 &        58.0 &       55.4 &     77.5 &       24.1 & 59.0\\
                           & MOCO v1  &  56.6 &   58.3 &        52.9 &       37.5 &     71.7 &       19.2 & 49.4\\
                           & MOCO v2  &  73.2 &   65.7 &        66.9 &       51.9 &     81.0 &       26.2 & 60.8\\
                           & Supervised (IN1k)  &  \textbf{80.6} &   \textbf{71.6} &        \textbf{68.9} &       \textbf{66.6} &     \textbf{81.6} &       \textbf{27.3} & \textbf{66.1}\\
                           
\bottomrule
\end{tabular}
\end{footnotesize}
    
        \end{center}
\end{table}

In Table~\ref{tab:ssl2} we report the results on 4 datasets to compare the DINO SSL pre-training to a supervised pre-training on ImageNet21k on a larger architecture that uses Layer Normalization (VIT Base).
The results confirm that, in most cases, the gap between SSL and supervised pre-training is still quite large. We also report some instabilities of SHOT using the DINO pre-training (for example on VisDA dataset) that very rarely happened with supervised pre-training.

\begin{table}[h!]
\begin{center}
\caption{Target accuracy (\%) using different initialization weights of VIT Base after performing the following experiments: fine-tuning on the source domain (FT), FT+SCA and FT+SHOT.}
\vspace{0.5cm}
\label{tab:ssl2}

\begin{footnotesize}
\begin{tabular}{llcccc}
\toprule
\textbf{Experiment} & \textbf{Pre-train} & \textbf{O31} & \textbf{MO31} &  \textbf{Visda} &  \textbf{O.Home} \\
\midrule
\multirow{2}{*}{FT} & DINO     &  80.7 &   72.4 &        65.3 &       63.6 \\
                    & Supervised (IN21k)  &  \textbf{89.7} &   \textbf{86.9} &        \textbf{77.9} &       \textbf{79.9} \\
\midrule
\multirow{2}{*}{FT+SCA}    & DINO     &  85.2 &   81.1 &        73.6 &       67.9 \\
                           & Supervised (IN21k) &  \textbf{92.7} &   \textbf{91.9} &        \textbf{82.6} &       \textbf{82.5} \\
\midrule
\multirow{2}{*}{FT+SHOT}   & DINO    &  87.5 &   75.8 &     56.9 &       74.2 \\
                           & Supervised (IN21k)  &  \textbf{93.7} &   \textbf{93.9} &        \textbf{87.5} &       \textbf{84.4} \\
                           
\bottomrule
\end{tabular}
\end{footnotesize}
    
        \end{center}
\end{table}

\end{document}